\pdfoutput=1
\documentclass[11pt]{article}
\usepackage[utf8]{inputenc}
\usepackage{times}
\usepackage{longtable}
\usepackage{algorithm}
\usepackage[noend]{algorithmic}
\usepackage{nicefrac}
\usepackage{booktabs}
\usepackage{footmisc}
\usepackage[shortlabels]{enumitem}
\usepackage[margin=1in]{geometry}
\usepackage{comment}
\usepackage{booktabs,tabularx}
\usepackage{multirow}
\usepackage{textcomp}
\usepackage[numbers]{natbib}
\usepackage{graphicx}
\usepackage{color}
\usepackage[hidelinks,colorlinks=true,linkcolor=blue,citecolor=blue]{hyperref}
\usepackage{wrapfig}
\usepackage{amssymb,amsmath,amsthm}
\usepackage[capitalise,noabbrev]{cleveref}

\usepackage{array}
\usepackage{url}
\usepackage[strict]{changepage}
\usepackage{makecell}
\usepackage{titlesec}
\usepackage{setspace}
\usepackage{bm}
\usepackage{mdframed}
\usepackage{bbm}
\usepackage{threeparttable}
\usepackage{subfigure}

\usepackage[T1]{fontenc}

\usepackage{titletoc}
\usepackage{longtable}
\usepackage{pifont} 
\usepackage{wrapfig}

\usepackage{fancyhdr} % for fancy headers and footers
\usepackage{xcolor} % for color definitions
\usepackage[most]{tcolorbox} % for colored boxes

\usepackage{caption}
\usepackage{subfigure}

\titleformat*{\subparagraph}{\itshape}

\newcommand{\yx}[1]{#1}
\newsavebox\actorsfigure

\title{\texttt{Sora}: A Review on Background, Technology, Limitations, and Opportunities of Large Vision Models}

\author{
{\bfseries Yixin Liu$^{1}$\footnotemark[1]} \quad
{\bfseries Kai Zhang$^{1}$\footnotemark[1]} \quad
{\bfseries Yuan Li$^{1}$\footnotemark[1]} \quad
{\bfseries Zhiling Yan$^{1}$\footnotemark[1]} \quad
{\bfseries Chujie Gao$^{1}$\footnotemark[1]} \\
{\bfseries Ruoxi Chen$^{1}$\footnotemark[1]} \quad
{\bfseries Zhengqing Yuan$^{1}$\footnotemark[1]} \quad
{\bfseries Yue Huang$^{1}$\footnotemark[1]} \quad
{\bfseries Hanchi Sun$^{1}$\footnotemark[1]} \\
{\bfseries Jianfeng Gao$^{2}$} \quad
{\bfseries Lifang He$^{1}$} \quad
{\bfseries Lichao Sun$^{1}$\footnotemark[2]} \\
\\
{\bfseries $^{1}$Lehigh University}\quad
{\bfseries $^{2}$Microsoft Research}\quad
\\
}

\def \hhead#1{\noindent\textbf{#1}}

\date{}

\begin{document}

\maketitle  
\renewcommand{\thefootnote}{\fnsymbol{footnote}}
\footnotetext[1]{Equal contributions. The order was determined by rolling dice. Chujie, Ruoxi, Yuan, Yue, and Zhengqing are visiting students in the LAIR lab at Lehigh University. The GitHub link is \url{https://github.com/lichao-sun/SoraReview}}
\footnotetext[2]{Lichao Sun is co-corresponding author: \href{mailto:lis221@lehigh.edu}{\color{black}{lis221@lehigh.edu}}}

\begin{abstract}
{\;\;\;\;\;\;\;\;\;\;\;\;\;\;\;\;\;\;\;\;\;\;\;\;\;\textcolor{red}{\textbf{Note: This is not an official technical report from OpenAI. }}}\\
\texttt{Sora} is a text-to-video generative AI model, released by OpenAI in February 2024. 
The model is trained to generate videos of realistic or imaginative scenes from text instructions and show potential in simulating the physical world.
Based on public technical reports and reverse engineering, this paper presents a comprehensive review of the model's background, related technologies, applications, remaining challenges, and future directions of text-to-video AI models.
We first trace \texttt{Sora}'s development and investigate the underlying technologies used to build this ``world simulator''. 
Then, we describe in detail the applications and potential impact of \texttt{Sora} in multiple industries ranging from film-making and education to marketing.   
We discuss the main challenges and limitations that need to be addressed to widely deploy \texttt{Sora}, such as ensuring safe and unbiased video generation.
Lastly, we discuss the future development of \texttt{Sora} and video generation models in general, and how advancements in the field could enable new ways of human-AI interaction, boosting productivity and creativity of video generation.
\begin{figure}[h]
    \centering
    \includegraphics[width=0.57\linewidth]{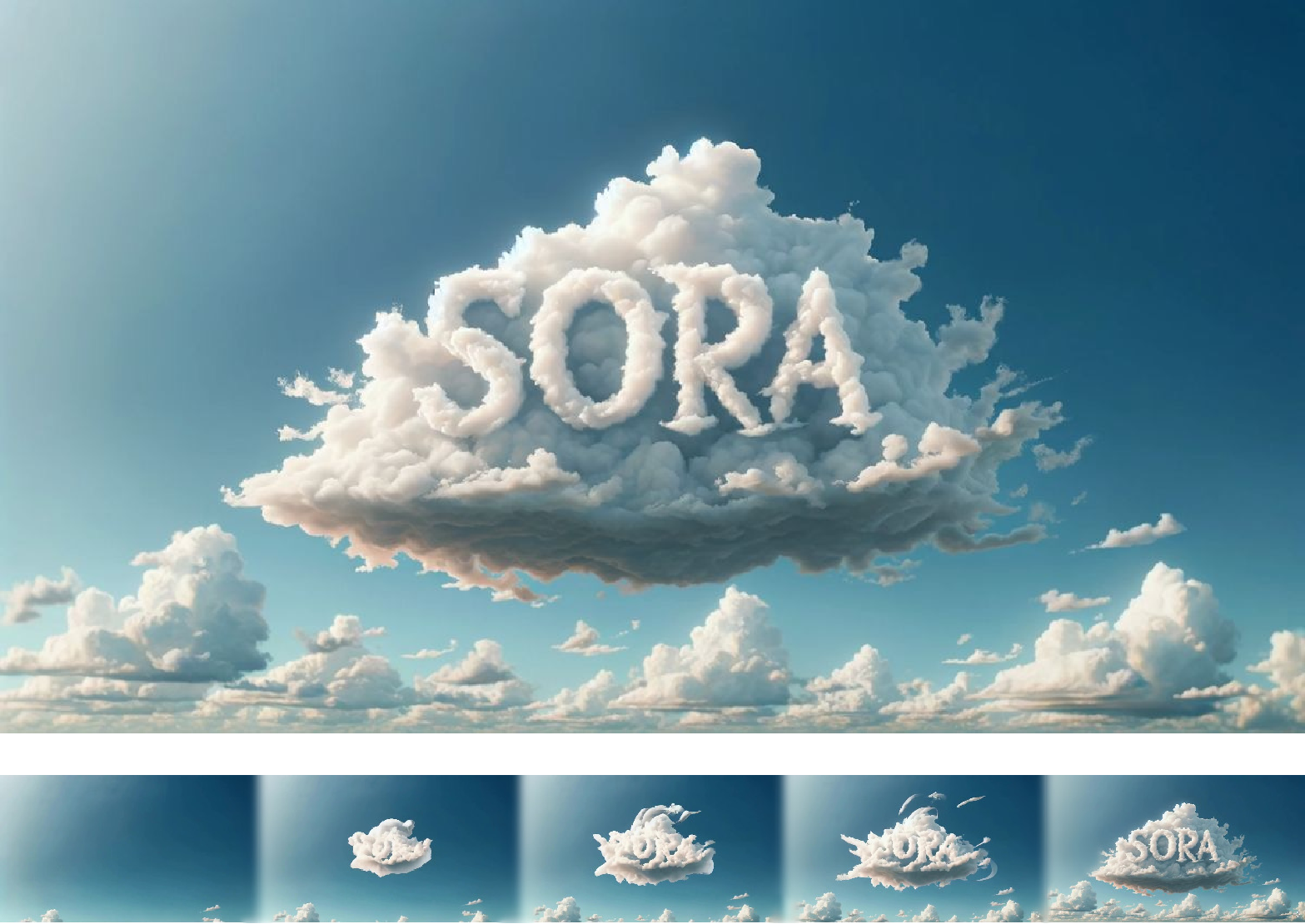}
    \caption{\texttt{Sora}: A Breakthrough in AI-Powered Vision Generation.}
    \label{fig:demo_example}
\end{figure}
\end{abstract}

\pagebreak

\begin{small}
\tableofcontents
\end{small}

\setlength{\parskip}{0.5em}

\pagebreak

\section{Introduction}

Since the release of ChatGPT in November 2022, the advent of AI technologies has marked a significant transformation, reshaping interactions and integrating deeply into various facets of daily life and industry~\cite{openai2022chatgpt,openai2023gpt4}. Building on this momentum,  OpenAI released, in February 2024, \texttt{Sora}, a text-to-video generative AI model that can generate videos of realistic or imaginative scenes from text prompts. 
%which stands at the forefront of advancements in the vision domain. 
Compared to previous video generation models, \texttt{Sora} is distinguished by its ability to produce up to 1-minute long videos with high quality while maintaining adherence to user's text instructions \cite{openai2024sora}. 
This progression of \texttt{Sora} is the embodiment of the long-standing AI research mission of equipping AI systems (or AI Agents) with the capability of understanding and interacting with the physical world in motion. This involves developing AI models that are capable of not only interpreting complex user instructions but also applying this understanding to solve real-world problems through dynamic and contextually rich simulations. 

\begin{figure}[ht]
    \centering
    \includegraphics[width=0.95\linewidth]{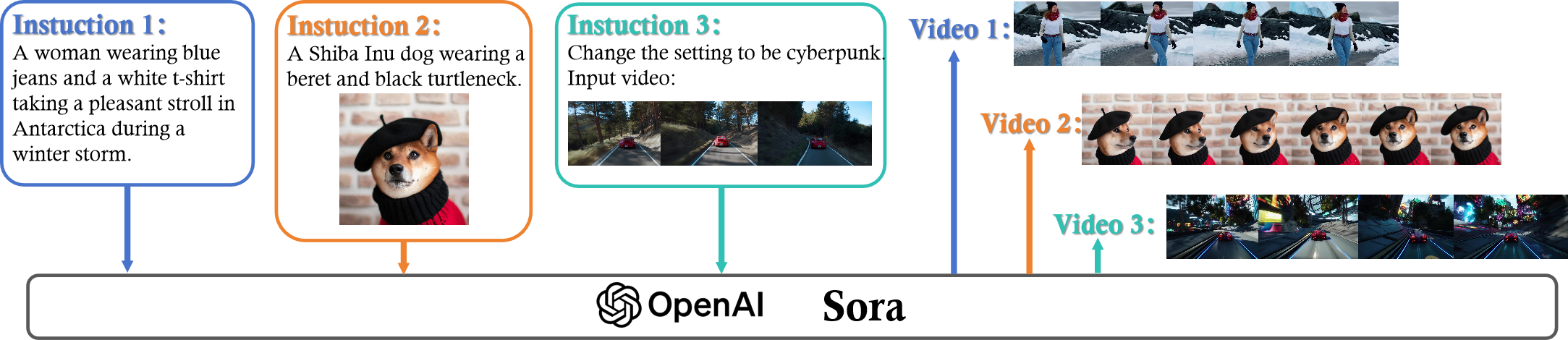}
    \caption{Examples of \texttt{Sora} in text-to-video generation. Text instructions are given to the OpenAI \texttt{Sora} model, and it generates three videos according to the instructions.}
    \label{fig:example}
    % \vspace{-10pt}
\end{figure}

\texttt{Sora} demonstrates a remarkable ability to accurately interpret and execute complex human instructions, as illustrated in Figure~\ref{fig:example}. The model can generate detailed scenes that include multiple characters that perform specific actions against intricate backgrounds. Researchers attribute \texttt{Sora}'s proficiency to not only processing user-generated textual prompts but also discerning the complicated interplay of elements within a scenario. One of the most striking aspects of \texttt{Sora} is its capacity for up to a minute-long video while maintaining high visual quality and compelling visual coherency. Unlike earlier models that can only generate short video clips, \texttt{Sora}'s minute-long video creation possesses a sense of progression and a visually consistent journey from its first frame to the last. In addition, \texttt{Sora}'s advancements are evident in its ability to produce extended video sequences with nuanced depictions of motion and interaction, overcoming the constraints of shorter clips and simpler visual renderings that characterized earlier video generation models. This capability represents a leap forward in AI-driven creative tools, allowing users to convert text narratives to rich visual stories. Overall, these advances show the potential of \texttt{Sora} as a \emph{world simulator} to provide nuanced insights into the physical and contextual dynamics of the depicted scenes. \cite{openai2024sora}. 

\hhead{Technology.} At the heart of \texttt{Sora} is a pre-trained \emph{diffusion transformer} \cite{peebles2023scalable}. Transformer models have proven scalable and effective for many natural language tasks. Similar to powerful large language models (LLMs) such as GPT-4, \texttt{Sora} can parse text and comprehend complex user instructions. %This ability forms the cornerstone of its remarkable translation of words into vivid, dynamic imagery. 
To make video generation computationally efficient, \texttt{Sora} employs \emph{spacetime latent patches} as its building blocks. Specifically, 
%a video is decomposed into smaller, more manageable segments.
\texttt{Sora} compresses a raw input video into a latent spacetime representation.
%each broken down into smaller patches that 
Then, a sequence of latent spacetime patches is extracted from the compressed video to
encapsulate both the visual appearance and motion dynamics over brief intervals. 
These patches, analogous to word tokens in language models, provide \texttt{Sora} with detailed \emph{visual phrases} to be used to construct videos. 
\texttt{Sora}'s text-to-video generation is performed by a diffusion transformer model. Starting with a frame filled with visual noise, the model iteratively denoises the image and introduces specific details according to the provided text prompt. In essence, the generated video emerges through a multi-step refinement process, with each step refining the video to be more aligned with the desired content and quality.

\hhead{Highlights of \texttt{Sora}.} \texttt{Sora}'s capabilities have profound implications in various aspects: % across various sectors:
\begin{itemize}[noitemsep,topsep=0pt]
    \item \textit{Improving simulation abilities}: Training \texttt{Sora} at scale is attributed to its remarkable ability to simulate various aspects of the physical world. Despite lacking explicit 3D modeling, \texttt{Sora} exhibits 3D consistency with dynamic camera motion and long-range coherence that includes object persistence and simulates simple interactions with the world. Moreover, \texttt{Sora} intriguingly simulates digital environments like Minecraft, controlled by a basic policy while maintaining visual fidelity. These emergent abilities suggest that scaling video models is effective in creating AI models to simulate the complexity of physical and digital worlds.
    \item \textit{Boosting creativity}: 
    %Artists, filmmakers, and designers face fewer constraints for visualizing their ideas.
    Imagine outlining a concept through text, whether a simple object or a full scene, and seeing a realistic or highly stylized video rendered within seconds.  \texttt{Sora} allows an accelerated design process for faster exploration and refinement of ideas, thus significantly boosting the creativity of artists, filmmakers, and designers.
    \item \textit{Driving educational innovations}:  Visual aids have long been integral to understanding important concepts in education. With \texttt{Sora}, educators can easily turn a class plan from text to videos
    %take written class plans and swiftly generate tailored videos 
    to captivate students' attention and improve learning efficiency. From scientific simulations to historical dramatizations, the possibilities are boundless.
    \item \textit{Enhancing Accessibility}: Enhancing accessibility in the visual domain is paramount. \texttt{Sora} offers an innovative solution by converting textual descriptions to visual content. This capability empowers all individuals, including those with visual impairments, to actively engage in content creation and interact with others in more effective ways. Consequently, it allows for a more inclusive environment where everyone has the opportunity to express his or her ideas through videos.
    \item \textit{Fostering emerging applications}: The applications of \texttt{Sora} are vast. For example, marketers might use it to create dynamic advertisements tailored to specific audience descriptions. Game developers might use it to generate customized visuals or even character actions from player narratives.
\end{itemize}

\hhead{Limitations and Opportunities.} While \texttt{Sora}'s achievements highlight significant advancements in AI, challenges remain. Depicting complex actions or capturing subtle facial expressions are among the areas where the model could be enhanced. In addition, ethical considerations such as mitigating biases in generated content and preventing harmful visual outputs underscore the importance of responsible usage by developers, researchers, and the broader community. Ensuring that \texttt{Sora}'s outputs are consistently safe and unbiased is a principal challenge.
The field of video generation is advancing swiftly, with academic and industry research teams making relentless strides. The advent of competing text-to-video models suggests that \texttt{Sora} may soon be part of a dynamic ecosystem. This collaborative and competitive environment fosters innovation, leading to improved video quality and new applications that help improve the productivity of workers and make people's lives more entertaining.

\hhead{Our Contributions.}
Based on published technical reports and our reverse engineering, this paper presents the first comprehensive review of \texttt{Sora}'s background, related technologies, emerging applications, current limitations, and future opportunities.

\vspace{-10pt}
\section{Background}
\vspace{-10pt}

\subsection{History}
\vspace{-10pt}

% Hanchi: VAE is earlier than GAN
In the realm of computer vision (CV), prior to the deep learning revolution, traditional image generation techniques relied on methods like texture synthesis~\cite{efros1999texture} and texture mapping~\cite{heckbert1986survey}, based on hand-crafted features. However, these methods were limited in their capacity to produce complex and vivid images. The introduction of Generative Adversarial Networks (GANs)~\cite{goodfellow2014generative} and Variational Autoencoders (VAEs)~\cite{kingma2013auto} marked a significant turning point due to its remarkable capabilities across various applications. Subsequent developments, such as flow models~\cite{dinh2014nice} and diffusion models~\cite{song2019DSM}, further enhanced image generation with greater detail and quality. The recent progress in Artificial Intelligence Generated Content (AIGC) technologies has democratized content creation, enabling users to generate desired content through simple textual instructions~\cite{cao2023comprehensive}.

\begin{figure}[t]
    \centering
    \includegraphics[width=0.95\linewidth]{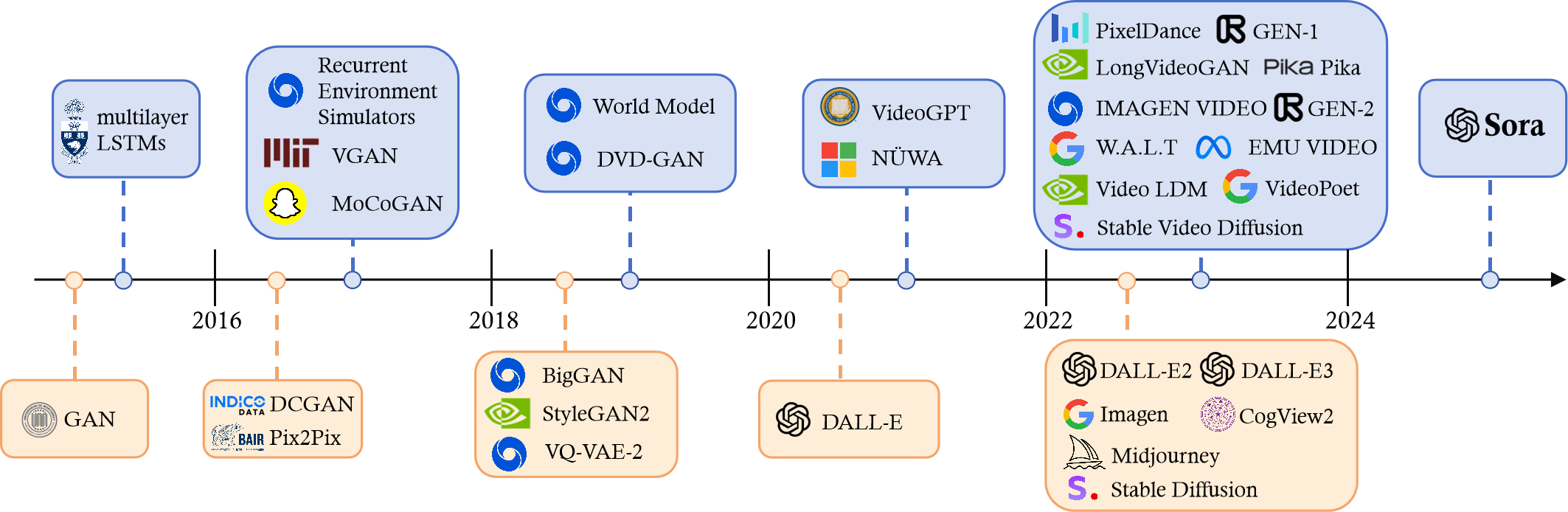}
    \caption{History of Generative AI in Vision Domain.}
    \label{fig: history}
\end{figure}

Over the past decade, the development of generative CV models has taken various routes, as shown in Figure~\ref{fig: history}. This landscape began to shift notably following the successful application of the transformer architecture~\cite{NIPS2017_transformer} in NLP, as demonstrated by BERT~\cite{devlin2018bert} and GPT~\cite{radford2018improving}. In CV, researchers take this concept even further by combining the transformer architecture with visual components, allowing it to be applied to downstream CV tasks, such as Vision Transformer (ViT)~\cite{dosovitskiy2020image} and Swin Transformer~\cite{liu2021swin}. Parallel to the transformer's success, diffusion models have also made significant strides in the fields of image and video generation~\cite{song2019DSM}. Diffusion models offer a mathematically sound framework for converting noise into images with U-Nets~\cite{ronneberger2015u}, where U-Nets facilitate this process by learning to predict and mitigate noise at each step. Since 2021, a paramount focus in AI has been on generative language and vision models that are capable of interpreting human instructions, known as multimodal models. For example, CLIP~\cite{clip2021} is a pioneering vision-language model that combines transformer architecture with visual elements, facilitating its training on vast datasets of text and images. By integrating visual and linguistic knowledge from the outset, CLIP can function as an image encoder within multimodal generation frameworks. Another notable example is Stable Diffusion~\cite{rombach2022high}, a versatile text-to-image AI model celebrated for its adaptability and ease of use. It employs transformer architecture and latent diffusion techniques to decode textual inputs and produce images of a wide array of styles, further illustrating the advancements in multimodal AI.

Following the release of ChatGPT in November 2022, we have witnessed the emergence of commercial text-to-image products in 2023, such as Stable Diffusion~\cite{rombach2022high}, Midjourney~\cite{midjourney2023}, DALL-E 3~\cite{betker2023improving}. These tools enable users to generate new images of high resolution and quality with simple text prompts, showcasing the potential of AI in creative image generation. However, transitioning from text-to-image to text-to-video is challenging due to the temporal complexity of videos. Despite numerous efforts in industry and academia, most existing video generation tools, such as Pika~\cite{pika2023} and Gen-2~\cite{gen22023}, are limited to producing only short video clips of a few seconds. In this context, \texttt{Sora} represents a significant breakthrough, akin to ChatGPT's impact in the NLP domain. \texttt{Sora} is the first model that is capable of generating videos up to one minute long based on human instructions, marking a milestone that profoundly influences research and development in generative AI. To facilitate easy access to the latest advancements in vision generation models, the most recent works have been compiled and provided in the Appendix and our GitHub. %This curated list aims to serve as a valuable resource, keeping readers informed about cutting-edge developments in the field and enabling them to explore these innovative models further. 

\subsection{Advanced Concepts}
\hhead{Scaling Laws for Vision Models.}~~ With scaling laws for LLMs, it is natural to ask whether the development of vision models follows similar scaling laws.
%signatures as LLMs so that the scaling up of vision models can catch up with LLMs. 
Recently, Zhai et al.~\cite{zhai2022scaling} have demonstrated that the performance-compute frontier for ViT models with enough training data roughly follows a (saturating) power law. Following them, Google Research~\cite{dehghani2023scaling} presented a recipe for highly efficient and stable training of a 22B-parameter ViT. Results show that great performance can be achieved using the frozen model to produce embeddings, and then training thin layers on top. \texttt{Sora}, as a large vision model (LVM), aligns with these scaling principles, uncovering several emergent abilities in text-to-video generation. This significant progression underscores the potential for LVMs to achieve advancements like those seen in LLMs.

\hhead{Emergent Abilities.}~~Emergent abilities in LLMs are sophisticated behaviors or functions that manifest at certain scales—often linked to the size of the model's parameters—that were not explicitly programmed or anticipated by their developers. These abilities are termed "emergent" because they emerge from the model's comprehensive training across varied datasets, coupled with its extensive parameter count. This combination enables the model to form connections and draw inferences that surpass mere pattern recognition or rote memorization. Typically, the emergence of these abilities cannot be straightforwardly predicted by extrapolating from the performance of smaller-scale models. While numerous LLMs, such as ChatGPT and GPT-4, exhibit emergent abilities, vision models demonstrating comparable capabilities have been scarce until the advent of Sora. According to \texttt{Sora}'s technical report, it is the first vision model to exhibit confirmed emergent abilities, marking a significant milestone in the field of computer vision.

In addition to its emergent abilities, \texttt{Sora} exhibits other notable capabilities, including instruction following, visual prompt engineering, and video understanding. These aspects of \texttt{Sora}'s functionality represent significant advancements in the vision domain and will be explored and discussed in the rest sections.

\vspace{-10pt}
\section{Technology}
\vspace{-10pt}
\subsection{Overview of \texttt{Sora}} % Hanchi
\vspace{-10pt}

\begin{figure}[b]
    \centering
    \includegraphics[width=1.0\linewidth]{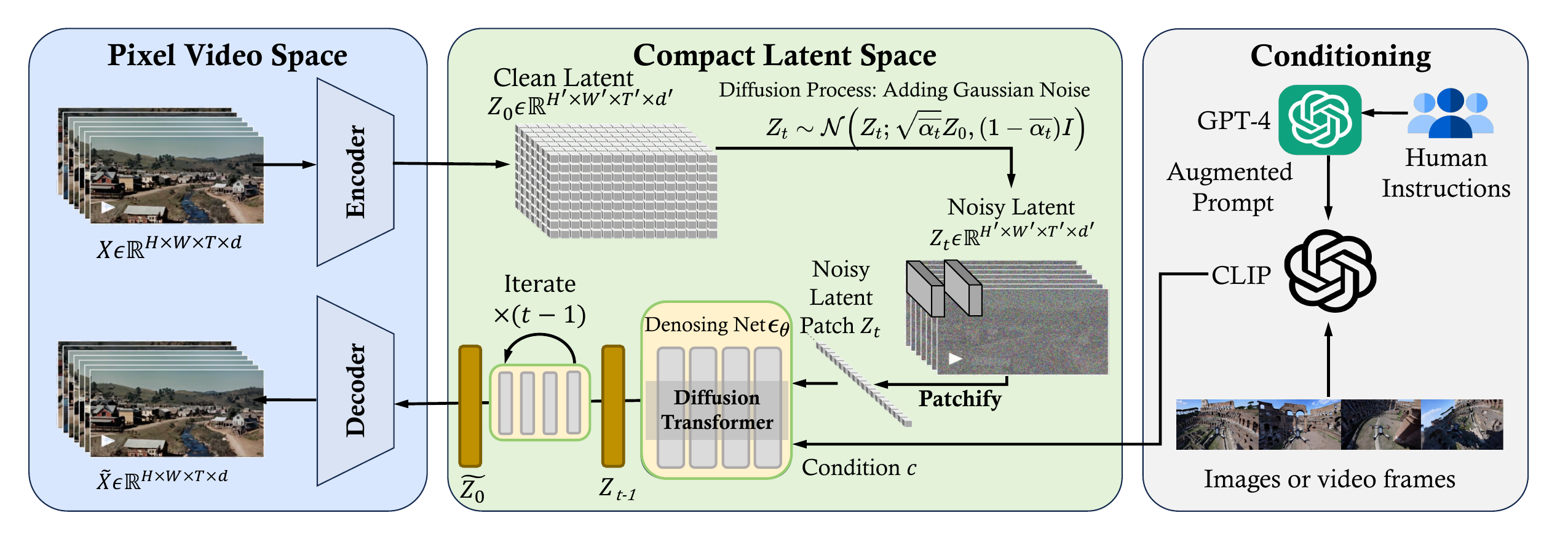}
    \caption{\textbf{Reverse Engineering}: Overview of \texttt{Sora} framework}
    \label{fig:sora_framework}
\end{figure}

In the core essence, \texttt{Sora} is a diffusion transformer \cite{peebles2023scalable} with flexible sampling dimensions as shown in Figure \ref{fig:sora_framework}. It has three parts: (1) A time-space compressor first maps the original video into latent space. (2) A ViT then processes the tokenized latent representation and outputs the denoised latent representation. (3) A CLIP-like \cite{radford2021learning} conditioning mechanism receives LLM-augmented user instructions and potentially visual prompts to guide the diffusion model to generate styled or themed videos. After many denoising steps, the latent representation of the generated video is obtained and then mapped back to pixel space with the corresponding decoder. In this section, we aim to reverse engineer the technology used by \texttt{Sora} and discuss a wide range of related works. 

\vspace{-10pt}
\subsection{Data Pre-processing}
\vspace{-10pt}
\subsubsection{Variable Durations, Resolutions, Aspect Ratios} \label{sec:vary_data}
\vspace{-10pt}

\begin{figure}[h]
    \centering
    \includegraphics[width=0.9\linewidth]{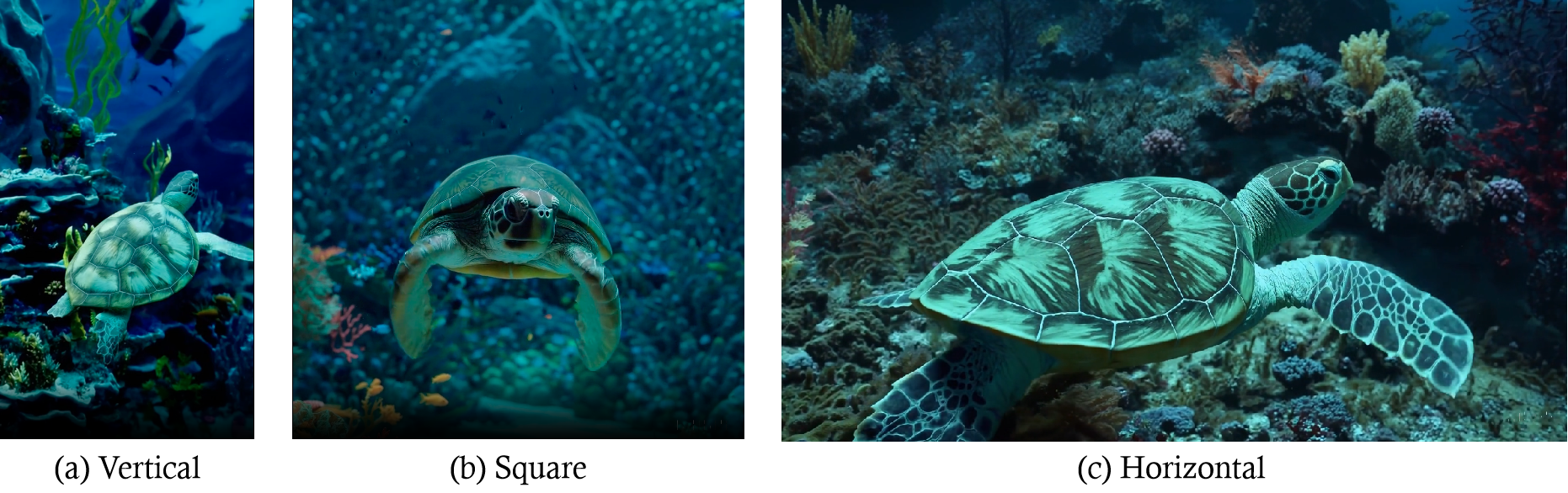}
    \caption{\texttt{Sora} can generate images in flexible sizes or resolutions ranging from 1920x1080p to 1080x1920p and anything in between. }
    \label{fig:flexible-generation-sizes}
\end{figure}

One distinguishing feature of \texttt{Sora} is its ability to train on, understand, and generate videos and images at their native sizes \cite{openai2024sora} as illustrated in Figure \ref{fig:flexible-generation-sizes}. Traditional methods often resize, crop, or adjust the aspect ratios of videos to fit a uniform standard—typically short clips with square frames at fixed low resolutions \cite{blattmann2023stable}\cite{singer2022makeavideo}\cite{ho2022imagen}. Those samples are often generated at a wider temporal stride and rely on separately trained frame-insertion and resolution-rendering models as the final step, creating inconsistency across the video. Utilizing the diffusion transformer architecture \cite{peebles2023scalable} (see Section \ref{sec:space_time_patch}), \texttt{Sora} is the first model to embrace the diversity of visual data and can sample in a wide array of video and image formats, ranging from widescreen 1920x1080p videos to vertical 1080x1920p videos and everything in between without compromising their original dimensions. 

\begin{wrapfigure}{r}{0.5\textwidth} 
\vspace{-10pt}
    \centering
    \includegraphics[width=\linewidth]{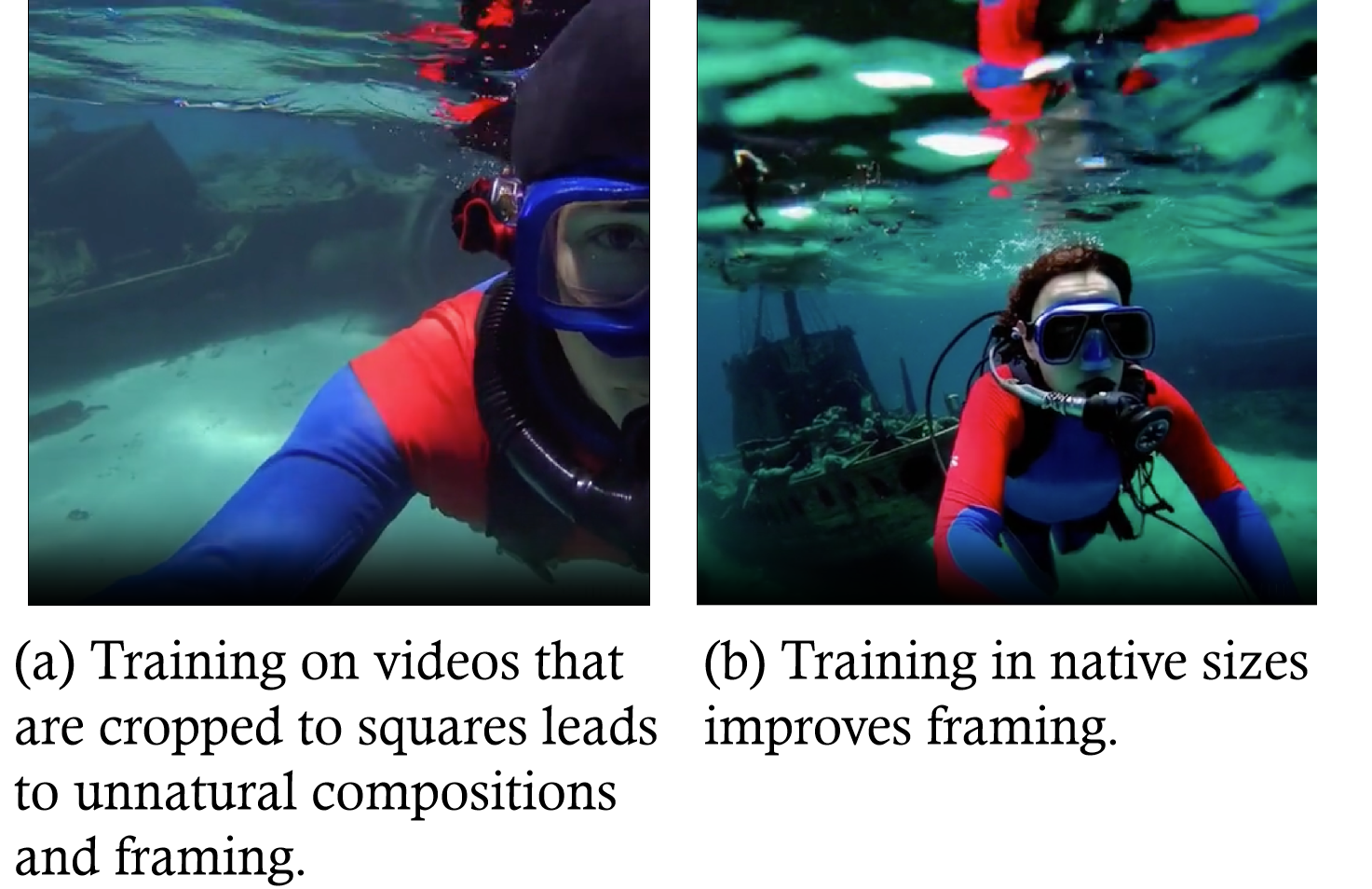}
    \vspace{-20pt}
    \caption{A comparison between \texttt{Sora} (right) and a modified version of the model (left), which crops videos to square shapes—a common practice in model training—highlights the advantages.}
    \vspace{-10pt}
    \label{fig:native-sizes-vs-cropped}
\end{wrapfigure}
Training on data in their native sizes significantly improves composition and framing in the generated videos. Empirical findings suggest that by maintaining the original aspect ratios, \texttt{Sora} achieves a more natural and coherent visual narrative. The comparison between \texttt{Sora} and a model trained on uniformly cropped square videos demonstrates a clear advantage as shown in Figure \ref{fig:native-sizes-vs-cropped}. Videos produced by \texttt{Sora} exhibit better framing, ensuring subjects are fully captured in the scene, as opposed to the sometimes truncated views resulting from square cropping.

This nuanced understanding and preservation of original video and image characteristics mark a significant advancement in the field of generative models. \texttt{Sora}'s approach not only showcases the potential for more authentic and engaging video generation but also highlights the importance of diversity in training data for achieving high-quality results in generative AI. The training approach of \texttt{Sora} aligns with the core tenet of Richard Sutton's \textsc{The Bitter Lesson\cite{Sutton2019BitterLesson}}, which states that leveraging computation over human-designed features leads to more effective and flexible AI systems. Just as the original design of diffusion transformers seeks simplicity and scalability \cite{xie2024twitter}, Sora's strategy of training on data at their native sizes eschews traditional AI reliance on human-derived abstractions, favoring instead a generalist method that scales with computational power. In the rest of this section, we try to reverse engineer the architecture design of \texttt{Sora} and discuss related technologies to achieve this amazing feature. 
\vspace{-10pt}
\subsubsection{Unified Visual Representation}
\vspace{-10pt}

To effectively process diverse visual inputs including images and videos with varying durations, resolutions, and aspect ratios, a crucial approach involves transforming all forms of visual data into a unified representation, which facilitates the large-scale training of generative models. Specifically, \texttt{Sora} patchifies videos by initially compressing videos into a lower-dimensional latent space, followed by decomposing the representation into spacetime patches. However, \texttt{Sora}'s technical report \cite{openai2024sora} merely presents a high-level idea, making reproduction challenging for the research community. In this section, we try to reverse-engineer the potential ingredients and technical pathways. Additionally, we will discuss viable alternatives that could replicate \texttt{Sora}'s functionalities, drawing upon insights from existing literature. 

\begin{figure}[h]
    \centering
    \includegraphics[width=.95\linewidth]{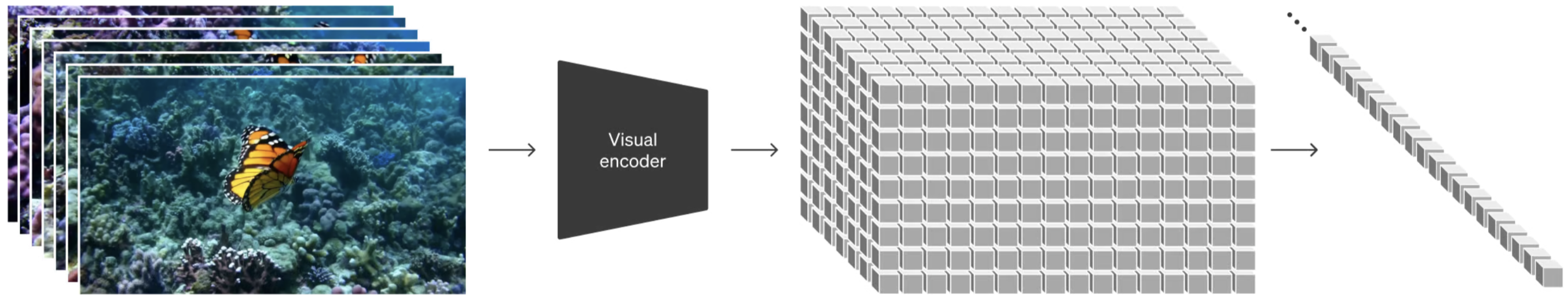}
    \caption{At a high level, \texttt{Sora} turns videos into patches by first compressing videos into a lower-dimensional latent space, and subsequently decomposing the representation into spacetime patches. Source: \texttt{Sora}'s technical report \cite{openai2024sora}.}
    \label{fig:compression}
    \vspace{-10pt}
\end{figure}

\vspace{-10pt}
\subsubsection{Video Compression Network} 
\label{sec. compress-net}
\vspace{-10pt}

\begin{wrapfigure}{r}{0.4\textwidth} 
% \vspace{-20pt}
% \vspace{-230pt}
    \centering
    \includegraphics[width=\linewidth]{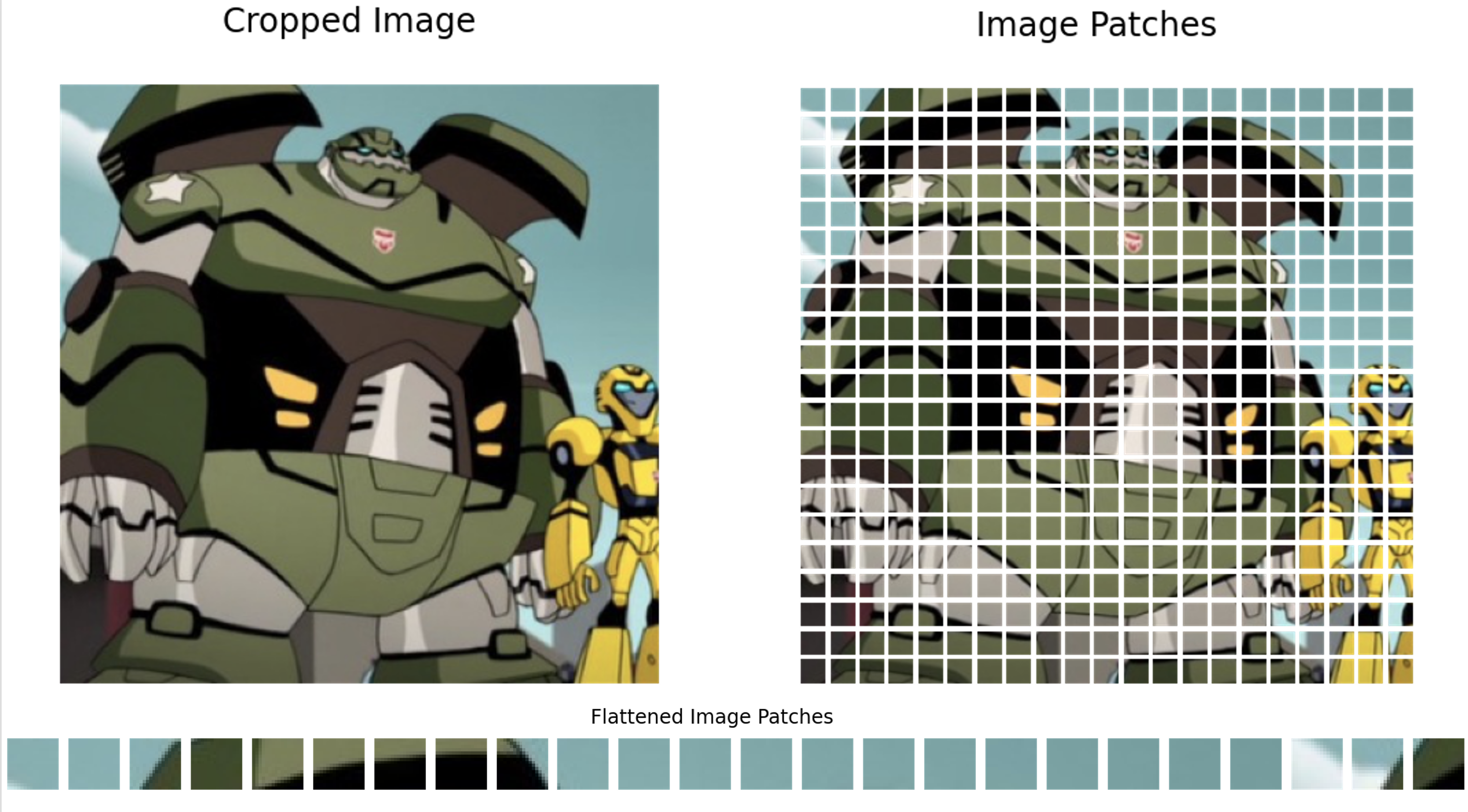}
    \caption{ViT splits an image into fixed-size patches, linearly embeds each of them, adds position embeddings, and feeds the resulting sequence of vectors to a standard Transformer encoder.}
    \label{fig:vit}
% \vspace{-5pt}
\end{wrapfigure}
\begin{figure}[h]
\centering 
\subfigure[\textbf{Additional temporal layer}. A pre-trained LDM is turned into a video generator by inserting temporal layers that learn to align frames into temporally consistent sequences. During optimization, the image backbone \(\theta\) remains fixed and only the parameters \(\phi\) of the temporal layers \(l_{\phi}^{i}\) are trained. 
]{
    \begin{minipage}{0.47\linewidth}
    \centering   
    \includegraphics[width=\linewidth]{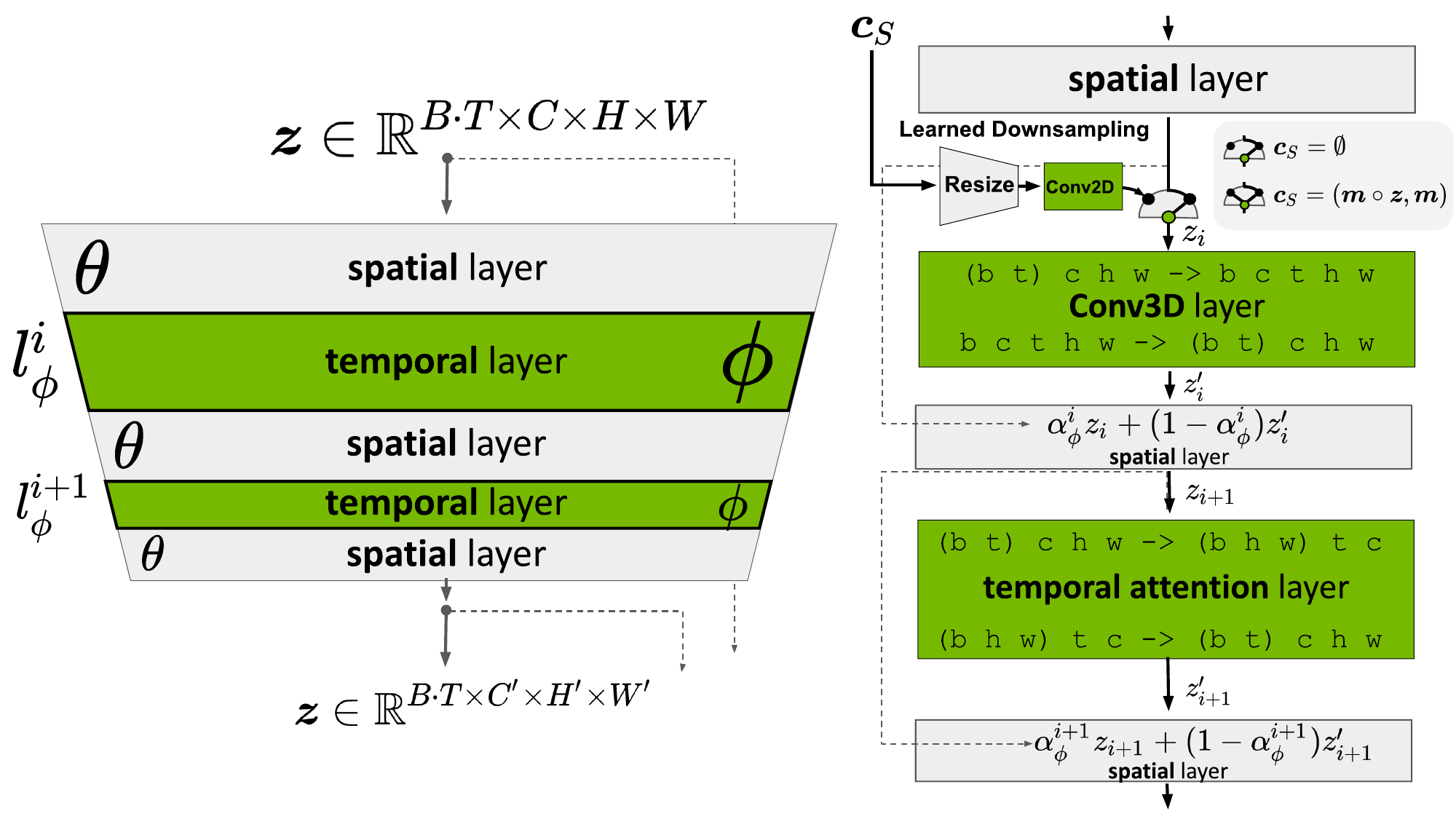}
    \end{minipage}
}
\hfill % Adds a space between the subfigures. Remove or adjust as needed.
\subfigure[\textbf{Video LDM stack.} Video LDM first generates sparse key frames and then temporally interpolates twice with the same latent diffusion models to achieve a high frame rate. Finally, the latent video is decoded to pixel space, and optionally, a video upsampler diffusion model is applied.]{
    \begin{minipage}{0.47\linewidth}
    \centering   
    \includegraphics[width=\linewidth]{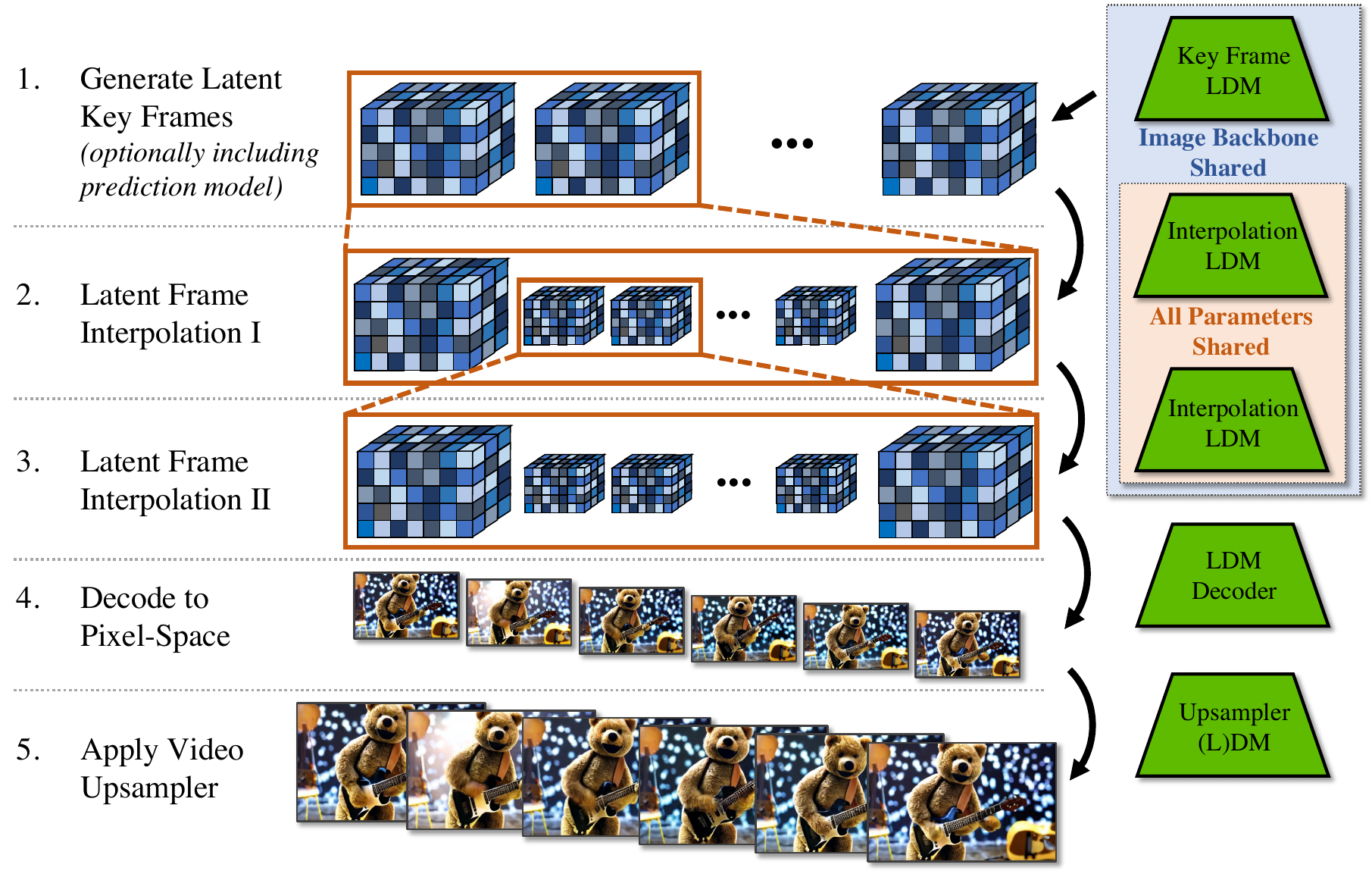}
    \end{minipage}
}
\caption{The overall framework of Video LDM. Source: Video LDM \cite{blattmann2023align}.}
\label{fig:VLDM}   
\end{figure}

\texttt{Sora}'s video compression network (or visual encoder) aims to reduce the dimensionality of input data, especially a raw video, and output a latent representation that is compressed both temporally and spatially as shown in Figure \ref{fig:compression}. According to the references in the technical report, the compression network is built upon VAE or Vector Quantised-VAE (VQ-VAE) \cite{van2017neural}. However, it is challenging for VAE to map visual data of any size to a unified and fixed-sized latent space if resizing and cropping are not used as mentioned in the technical report. We summarize two distinct implementations 
%but in a hierarchy 
to address this issue:

\hhead{Spatial-patch Compression.}~~This involves transforming video frames into fixed-size patches, akin to the methodologies employed in ViT \cite{dosovitskiy2020image} and MAE \cite{he2022masked} (see Figure \ref{fig:vit}), before encoding them into a latent space. This approach is particularly effective for accommodating videos of varying resolutions and aspect ratios, as it encodes entire frames through the processing of individual patches. Subsequently, these spatial tokens are organized in a temporal sequence to create a spatial-temporal latent representation. This technique highlights several critical considerations: \textit{\underline{Temporal dimension variability}} -- given the varying durations of training videos, the temporal dimension of the latent space representation cannot be fixed. To address this, one can either sample a specific number of frames (padding or temporal interpolation \cite{ge2023preserve} may be needed for much shorter videos) or define a universally extended (super long) input length for subsequent processing (more details are described in Section \ref{sec:space_time_patch}); \textit{\underline{Utilization of pre-trained visual encoders}} -- for processing videos of high resolution, leveraging existing pre-trained visual encoders, such as the VAE encoder from Stable Diffusion \cite{rombach2022high}, is advisable for most researchers while \texttt{Sora}'s team is expected to train their own compression network with a decoder (the video generator) from scratch via the manner employed in training latent diffusion models \cite{rombach2022high, sauer2023adversarial, blattmann2023align}. These encoders can efficiently compress large-size patches (e.g., $256 \times 256$), facilitating the management of large-scale data; \textit{\underline{Temporal information aggregation}} -- since this method primarily focuses on spatial patch compression, it necessitates an additional mechanism for aggregating temporal information within the model. This aspect is crucial for capturing dynamic changes over time and is further elaborated in subsequent sections (see details in Section \ref{sec:dit} and Figure \ref{fig:VLDM}).
    % via spatial then temporal autoregressive

\begin{figure}[htbp]
    \centering 
    % \subfigure[Spatial patchification simply samples $n_t$ frames, and embed each 2D frame independently following ViT.]{   
    \begin{minipage}{9cm}
    \centering   
    \includegraphics[height=3.2cm]{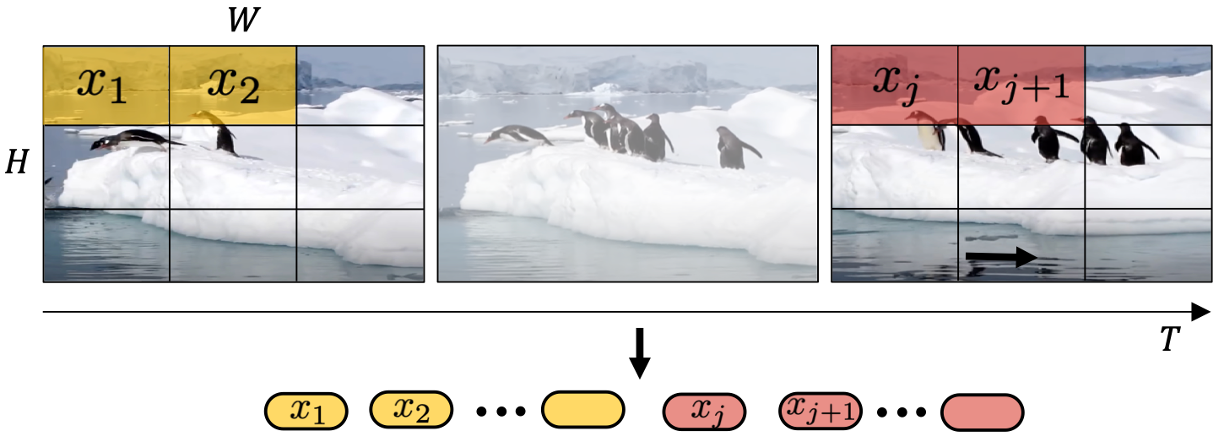}
    \end{minipage}
    % }
    % \subfigure[Spatial-temporal patchification extracts and linearly embed non-overlapping or overlapping tubelets that span the spatio-temporal input volume.]{ 
    \begin{minipage}{7cm}
    \centering   
    \includegraphics[height=4cm]{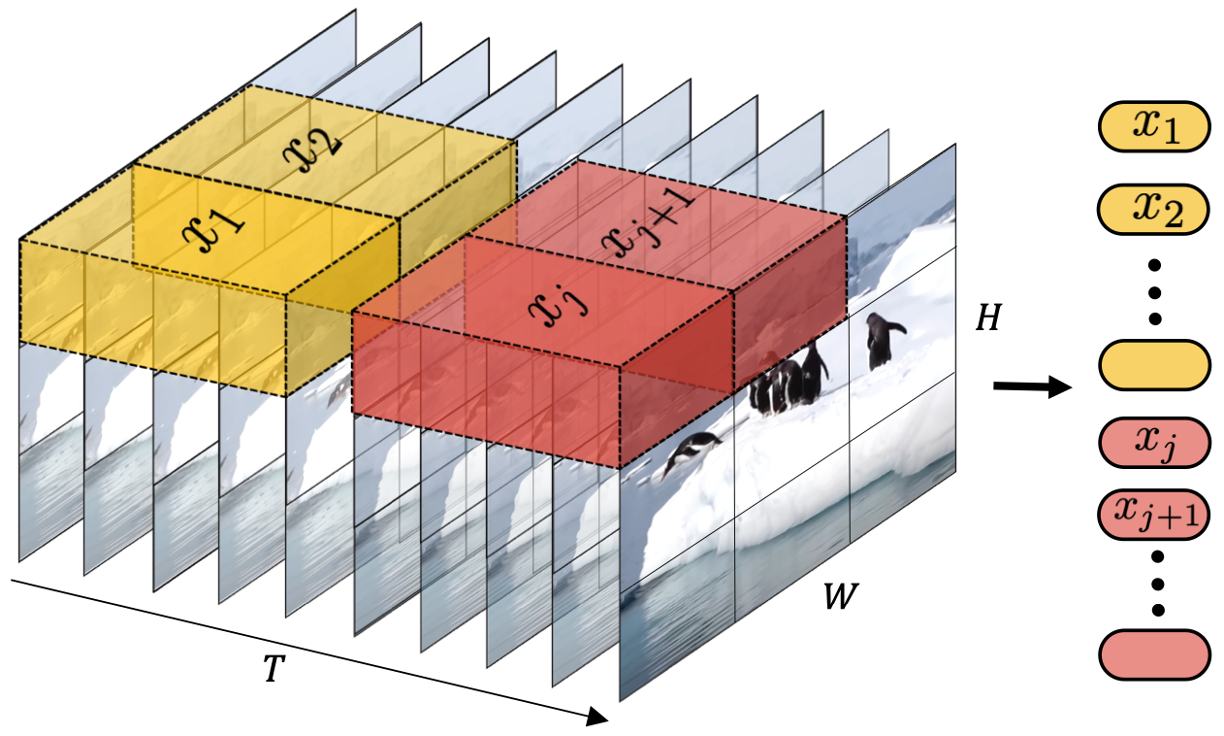}
    \end{minipage}
    % }
    \caption{Comparison between different patchification for video compression. Source: ViViT \cite{arnab2021vivit}. (\textbf{Left}) Spatial patchification simply samples $n_t$ frames and embeds each 2D frame independently following ViT. (\textbf{Right}) Spatial-temporal patchification extracts and linearly embeds non-overlapping or overlapping tubelets that span the spatiotemporal input volume.}    
    \label{fig:video_clip_embedding}   
    \end{figure}

\hhead{Spatial-temporal-patch Compression.}~~This technique is designed to encapsulate both spatial and temporal dimensions of video data, offering a comprehensive representation. This technique extends beyond merely analyzing static frames by considering the movement and changes across frames, thereby capturing the video's dynamic aspects. The utilization of 3D convolution emerges as a straightforward and potent method for achieving this integration \cite{ryoo2021tokenlearner}. The graphical illustration and the comparison against pure spatial-pachifying are depicted in Figure \ref{fig:video_clip_embedding}. Similar to spatial-patch compression, employing spatial-temporal-patch compression with predetermined convolution kernel parameters -- such as fixed kernel sizes, strides, and output channels -- results in variations in the dimensions of the latent space due to the differing characteristics of video inputs. This variability is primarily driven by the diverse durations and resolutions of the videos being processed. To mitigate this challenge, the approaches adopted for spatial patchification are equally applicable and effective in this context.

In summary, we reverse engineer the two patch-level compression approaches based on VAE or its variant like VQ-VQE because operations on patches are more flexible to process different types of videos. Since \texttt{Sora} aims to generate high-fidelity videos, a large patch size or kernel size is used for efficient compression. Here, we expect that fixed-size patches are used for simplicity, scalability, and training stability. But varying-size patches could also be used \cite{beyer2023flexivit} to make the dimension of the whole frames or videos in latent space consistent. However, it may result in invalid positional encoding, and cause challenges for the decoder to generate videos with varying-size latent patches.

\vspace{-10pt}
\subsubsection{Spacetime Latent Patches} \label{sec:space_time_patch}
\vspace{-10pt}

There is a pivotal concern remaining in the compression network part: How to handle the variability in latent space dimensions (i.e., the number of latent feature chunks or patches from different video types) before feeding patches into the input layers of the diffusion transformer. Here, we discuss several solutions. 
%the challenge in detail and its solutions.

\begin{figure}[t]
    \centering
    \includegraphics[width=.9\linewidth]{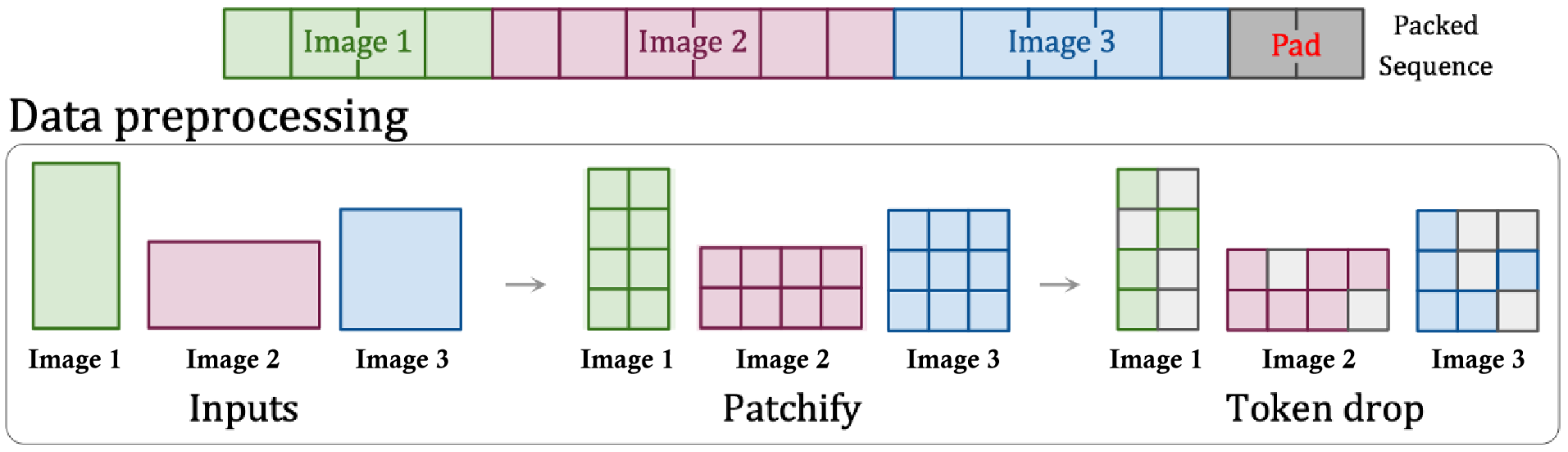}
    \caption{Patch packing enables variable resolution images or videos with preserved aspect ratio.6 Token dropping somehow could be treated as data augmentation. Source: NaViT \cite{dehghani2024patch}.}
    \label{fig:pnp_seq}
\end{figure}

Based on \texttt{Sora}'s technical report and the corresponding references, \textbf{patch n' pack (PNP)} \cite{dehghani2024patch} is likely the solution. %with high probability. 
PNP packs multiple patches from different images in a single sequence as shown in Figure \ref{fig:pnp_seq}. This method is inspired by example packing used in natural language processing \cite{krell2021efficient} that accommodates efficient training on variable length inputs by dropping tokens. Here the patchification and token embedding steps need to be completed in the compression network, but \texttt{Sora} may further patchify the latent for transformer token as Diffusion Transformer does \cite{peebles2023scalable}. 
Regardless there is a second-round patchification or not,
% We tend to believe there is no second-round patchification since the dimension of visual patches is not too large to be processed.
we need to address two concerns, how to pack those tokens in a compact manner and how to control which tokens should be dropped. For the first concern, a simple greedy approach is used which adds examples to the first sequence with enough remaining space. Once no more example can fit, sequences are filled with padding tokens, yielding the fixed sequence lengths needed for batched operations. Such a simple packing algorithm can lead to significant padding, depending on the distribution of the length of inputs. On the other hand, we can control the resolutions and frames we sample to ensure efficient packing by tuning the sequence length and limiting padding. For the second concern, an intuitive approach is to drop the similar tokens \cite{yin2022vit, bolya2022token, he2022masked, fayyaz2022adaptive} or, like PNP, apply dropping rate schedulers. However, it is worth noting that \textbf{\textit{3D Consistency}} is one of the nice properties of \texttt{Sora}. Dropping tokens may ignore fine-grained details during training. Thus, we believe that OpenAI is likely to use a super long context window and pack all tokens from videos although doing so is computationally expensive e.g., the multi-head attention \cite{vaswani2017attention, bertasius2021space} operator exhibits quadratic cost in sequence length. 
% (see detailed discussion in Section \ref{}). 
Specifically, spacetime latent patches from a long-duration video can be packed in one sequence while the ones from several short-duration videos are concatenated in the other sequence. 

% spatial and temporal position encoding (in modeling)

\vspace{-10pt}
\subsubsection{Discussion}
\vspace{-10pt}

We discuss two technical solutions to data pre-processing that \texttt{Sora} may use. Both solutions are performed at the patch level due to the characteristics of flexibility and scalability for modeling. Different from previous approaches where videos are resized, cropped, or trimmed to a standard size, \texttt{Sora} trains on data at its native size. Although there are several benefits (see detailed analysis in Section \ref{sec:vary_data}), it brings some technical challenges, among which one of the most significant is that neural networks cannot inherently process visual data of variable durations, resolutions, and aspect ratios. Through reverse engineering, we believe that \texttt{Sora} firstly compresses visual patches into low-dimensional latent representations, and arranges such latent patches or further patchified latent patches in a sequence, then injects noise into these latent patches before feeding them to the input layer of diffusion transformer. Spatial-temporal patchification is adopted by \texttt{Sora} because it is simple to implement, and it can effectively reduce the context length with high-information-density tokens and decrease the complexity of subsequent modeling of temporal information. 

% attention complexity, for research community -- use pre-trained models (how to adapt), reduce required resources, token drop

To the research community, we recommend using cost-efficient alternative solutions for video compression and representation, including utilizing pre-trained checkpoints (e.g., compression network) \cite{yu2023language}, shortening the context window, using light-weight modeling mechanisms such as (grouped) multi-query attention \cite{shazeer2019fast, ainslie2023gqa} or efficient architectures (e.g. Mamba \cite{gu2023mamba}), downsampling data and dropping tokens if necessary. The trade-off between effectiveness and efficiency for video modeling is an important research topic to be explored. 

\begin{figure}[t]
    \centering
    \includegraphics[width=.95\linewidth]{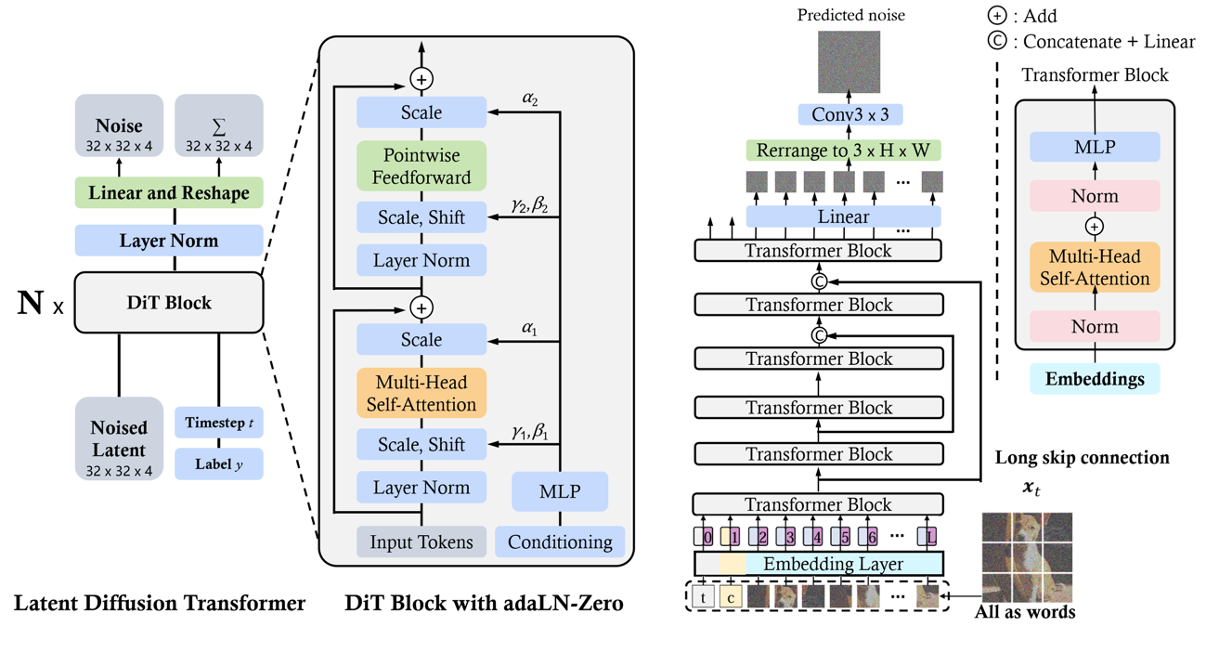}
    % \vspace{-10pt}
    \caption{The overall framework of DiT (left) and U-ViT (right)} \label{fig:overall-dit-uvit}.
    % \vspace{-10pt}
\end{figure}
% \vspace{-10pt}
\subsection{Modeling}
% \vspace{-10pt}
\subsubsection{Diffusion Transformer} \label{sec:dit}
% \vspace{-10pt}

% \vspace{-10pt}

\hhead{Image Diffusion Transformer.} Traditional diffusion models \cite{sohl-dickstein2015deep,ho2020denoising,song2020score} mainly leverage convolutional U-Nets that include downsampling and upsampling blocks for the denoising network backbone. However, recent studies show that the U-Net architecture is not crucial to the good performance of the diffusion model. By incorporating a more flexible transformer architecture, the transformer-based diffusion models can use more training data and larger model parameters. Along this line, 
DiT \cite{peebles2023scalable} and U-ViT \cite{Bao2023AllAW} are among the first works to %holistically investigate the design space in 
employ vision transformers for latent diffusion models. 
As in ViT, DiT employs a multi-head self-attention layer and a pointwise feed-forward network interlaced with some layer norm and scaling layers. Moreover, as shown in Figure \ref{fig:overall-dit-uvit}, DiT incorporates conditioning via adaptive layer norm (AdaLN) with an additional MLP layer for zero-initializing, which initializes each residual block as an identity function and thus greatly stabilizes the training process. The scalability and flexibility of DiT is empirically validated. DiT becomes the new backbone for diffusion models.
%the baselines for future research.
%and showcases the scalability and flexibility of DiTs as a backbone for generative modeling with diffusion models. 
In U-ViT, as shown in Figure \ref{fig:overall-dit-uvit}, they treat all inputs, including time, condition, and noisy image patches, as tokens and propose long skip connections between the shallow and deep transformer layers. The results suggest that the downsampling and upsampling operators in CNN-based U-Net are not always necessary, and U-ViT achieves record-breaking FID scores in image and text-to-image generation.

\begin{figure}[thbp]
    \centering
    \includegraphics[width=.9\linewidth]{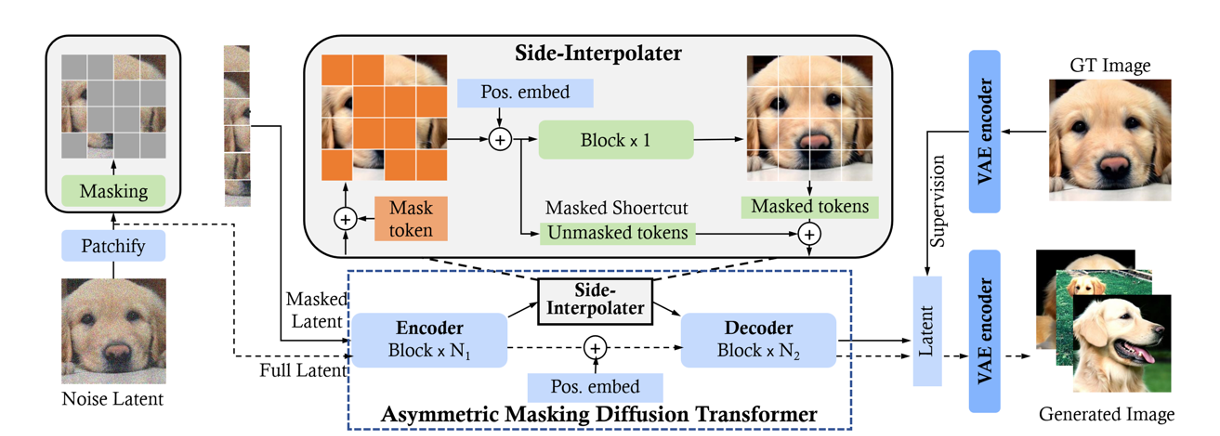}
    \caption{The overall framework of Masked Diffusion Transformer (MDT). A solid/dotted line indicates the training/inference process for each time step. Masking and side-interpolater are only used during training and are removed during inference.}
    \label{fig:overall-mdt}
\end{figure}

Like Masked AutoEncoder (MAE) \cite{he2022masked}, 
% in the vision transformer literature, 
Masked Diffusion Transformer (MDT) \cite{gao2023masked} incorporates mask latent modeling into the diffusion process to explicitly enhance contextual relation learning among object semantic parts in image synthesis. Specifically, as shown in Figure \ref{fig:overall-mdt}, MDT uses a side-interpolated for an additional masked token prediction task during training to boost the training efficiency and learn powerful context-aware positional embedding for inference. Compared to DiT \cite{peebles2023scalable}, MDT achieves better performance and faster learning speed. 
\yx{In MDTv2~\cite{gao2023masked_arxiv}, they further improve MDT with a more efficient macro network with some u-shape long-shortcuts in the encoder and dense shortcuts from encoder input in the decoder. Furthermore, they also leverage a bag of training strategies like the Adan optimizer~\cite{xie2022adan}, Min-SNR weighting~\cite{hang2023efficient}, dynamic masking ratio, and an improved power-cosine weighting for classifier-free guidance~\cite{ho2022classifier}. MDTv2 achieves a significant boost in synthesis performance and learning speed. }
Instead of using AdaLN (i.e., shifting and scaling) for time-conditioning modeling, Hatamizadeh et al.~\cite{hatamizadeh2023diffit} introduce Diffusion Vision Transformers (DiffiT), which uses a time-dependent self-attention (TMSA) module to model dynamic denoising behavior over sampling time steps. Besides, DiffiT uses two hybrid hierarchical architectures for efficient denoising in the pixel space and the latent space, respectively, and achieves new state-of-the-art results across various generation tasks. Overall, these studies show promising results in employing vision transformers for image latent diffusion, paving the way for future studies for other modalities. 

\hhead{Video Diffusion Transformer.} Building upon the foundational works in text-to-image (T2I) diffusion models, recent research has been focused on realizing the potential of diffusion transformers for text-to-video (T2V) generation tasks. Due to the temporal nature of videos, key challenges for applying DiTs in the video domain are: \textit{i) how to compress the video spatially and temporally to a latent space for efficient denoising}; \textit{ii) how to convert the compressed latent to patches and feed them to the transformer;} and \textit{iii) how to handle long-range temporal and spatial dependencies and ensure content consistency}. Please refer to Section \ref {sec. compress-net} for the first challenge. In this section, we focus our discussion on transformer-based denoising network architectures designed to operate in the spatially and temporally compressed latent space. \yx{We give a detailed review of the two important works (Imagen Video \cite{ho2022imagen} and Video LDM \cite{blattmann2023align}) shown in the OpenAI \texttt{Sora} technique report and also briefly introduce some of recent promising approaches. }
% Imagen Video \cite{ho2022imagen} is a cascade of video diffusion models, which consists of 7 sub-models that perform text-conditional video generation, spatial super-resolution, and temporal super-resolution. 

\begin{figure}[t]
    \centering
    \includegraphics[width=\linewidth]{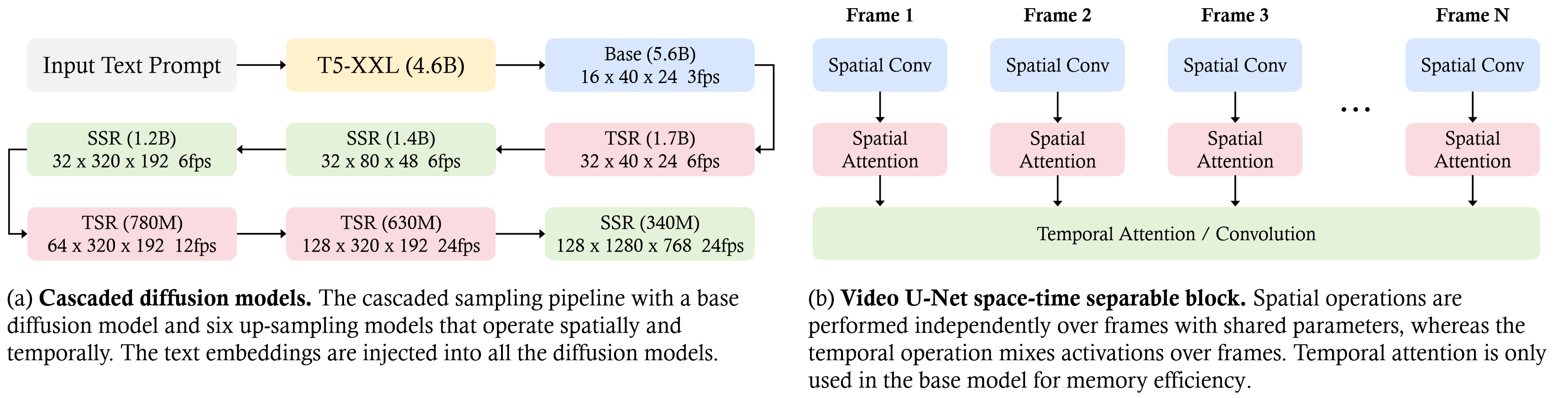}
    \caption{The overall framework of Imagen Video. Source: Imagen Video \cite{ho2022imagen}.}
    \label{fig:imagenV}
\end{figure}

Imagen Video \cite{ho2022imagen}, a text-to-video generation system developed by Google Research, utilizes a cascade of diffusion models, which consists of 7 sub-models that perform text-conditional video generation, spatial super-resolution, and temporal super-resolution, to transform textual prompts into high-definition videos. As shown in Figure \ref{fig:imagenV}, firstly, a frozen T5 text encoder generates contextual embeddings from the input text prompt. These embeddings are critical for aligning the generated video with the text prompt and are injected into all models in the cascade, in addition to the base model. Subsequently, the embedding is fed to the base model for low-resolution video generation, which is then refined by cascaded diffusion models to increase the resolution. The base video and super-resolution models use a 3D U-Net architecture in a spatial-temporal separable fashion. This architecture weaves temporal attention and convolution layers with spatial counterparts to efficiently capture inter-frame dependencies. It employs v-prediction parameterization for numerical stability and conditioning augmentation to facilitate parallel training across models. The process involves joint training on both images and videos, treating each image as a frame to leverage larger datasets, and using classifier-free guidance \cite{ho2022classifier} to enhance prompt fidelity. Progressive distillation \cite{salimans2022progressive} is applied to streamline the sampling process, significantly reducing the computational load while maintaining perceptual quality. Combining these methods and techniques allows Imagen Video to generate videos with not only high fidelity but also remarkable controllability, % and world knowledge, 
as demonstrated by its ability to produce diverse videos, text animations, and content in various artistic styles. 

Blattmann et al. \cite{blattmann2023align} propose to turn a 2D Latent Diffusion Model into a Video Latent Diffusion Model (Video LDM). They achieve this by adding some post-hoc temporal layers among the existing spatial layers into both the U-Net backbone and the VAE decoder that learns to align individual frames. These temporal layers are trained on encoded video data, while the spatial layers remain fixed, allowing the model to leverage large image datasets for pre-training. The LDM's decoder is fine-tuned for temporal consistency in pixel space and temporally aligning diffusion model upsamplers for enhanced spatial resolution. To generate very long videos, models are trained to predict a future frame given a number of context frames, allowing for classifier-free guidance during sampling. 
To achieve high temporal resolution, the video synthesis process is divided into key frame generation and interpolation between these keyframes. Following cascaded LDMs, a DM is used to further scale up the Video LDM outputs by four times, ensuring high spatial resolution while maintaining temporal consistency. This approach enables the generation of globally coherent long videos in a computationally efficient manner. Additionally, the authors demonstrate the ability to transform pre-trained image LDMs (e.g., Stable Diffusion) into text-to-video models by training only the temporal alignment layers, achieving video synthesis with resolutions up to $1280 \times 2048$px. 

\yx{Recently, there are also some other promising works that show superior performance in high-quality and temporal-consistent video generation. Similar to Imagen Video, LAVIE~\cite{wang2023lavie} proposes to transform a T2I model into a T2V model with cascaded LDMs, with two key insights: i) using simply temporal self-attention and RoPE~\cite{su2024roformer} can adequately capture temporal correlations inherent in video data; ii) joint image-video fine-tuning enable more stable training and avoid catastrophic forgetting. GenTron~\cite{chen2023gentron} adapt pre-trained class-conditioning DiTs to textual conditioning and explore different conditioning mechanisms, where they find that cross attention is a better conditioning method compared to previous adaLM-Zero for free-form textual conditioning. Furthermore, by designing motion-free masking in the inserted temporal self-attention layers, GenTron is able to do efficient image-video joint training with control on the temporal modeling ability in the video diffusion process. Window Attention Latent Transformer (W.A.L.T)~\cite{gupta2023photorealistic} proposes to unify the latent space of both videos and images with 3D causal CNN and a special coding strategy for the first frame. Moreover, W.A.L.T uses non-overlapping, window-restricted spatial and spatiotemporal attention to lower the computational cost and provide flexibility for image-video joint training. Latent Diffusion Transformer (Latte)~\cite{ma2024latte} conducts extensive experiments among four model variants from the perspective of decomposing spatial and temporal information in video data. Furthermore, Latte studies the effect of a wide range of training strategies and modules in a DiT-based video generation system, including conditioning mechanisms and temporal positional embedding. Instead of using separate compression for spatial and temporal dimensions in previous works, SnapVideo~\cite{menapace2024snap} proposes to extend EDM framework~\cite{karras2022elucidating} for high-dimensional video data. Furthermore, SnapVideo treats images as $T$ frames videos with infinite frame rates and introduces a variable frame-rate training procedure to bridge the modality gap. For scaling up to billion-parameters size, SnapVideo uses an efficient transformer-based architecture, FITs~\cite{chen2023fit}, as a scalable video generator, and a model cascade consisting of a first-stage lower-solution base model ($36\times64$px) and a high-solution upsampler ($288\times512$px) with corrupt training strategy. }

\vspace{-10pt}
\subsubsection{Discussion}
\vspace{-10pt}
\yx{
\hhead{One end-to-end model or cascade structure. }\texttt{Sora} can generate high-resolution videos. By reviewing existing works and our reverse engineering, we speculate that \texttt{Sora} also leverages cascade diffusion model architecture \cite{ho2022cascaded}, which is composed of a base model and many spatial-temporal refiner models. The attention modules are unlikely to be heavily used in the based diffusion model and low-resolution diffusion model, considering the high computation cost and limited performance gain of using attention machines in high-resolution cases. For spatial and temporal scene consistency, as previous works show that temporal consistency is more important than spatial consistency for video/scene generation, \texttt{Sora} is likely to leverage an efficient training strategy by using longer video (for temporal consistency) with lower resolution. Moreover, \texttt{Sora} is likely to use a $v$-parameterization diffusion model \cite{salimans2022progressive}, considering its superior performance among others that predict the original latent $x$ or the noise $\epsilon$. 

\hhead{On the compression of the latent encoder. }For training efficiency, most of the existing works leverage the pre-trained VAE encoder of Stable Diffusions \cite{rombach2021highresolution,podell2023sdxl}, a pre-trained 2D diffusion model, as an initialized model checkpoint. However, the encoder lacks the ability to compress temporally. Even though some works propose to only fine-tune the decoder for handling temporal information, the decoder's performance of dealing with video temporal data in the compressed latent space remains sub-optimal. Based on the technique report, our reverse engineering shows that, instead of using an existing pre-trained Image VAE encoder with some post-hoc temporal tuning, it is likely that \texttt{Sora} uses a spatial-temporal VAE encoder, jointly trained on image and video data, to compress the spatial and temporal information altogether. 
}
% on the training strategy to support video merging and backward and forward generation 
% TODO 

% \hhead{Bi-directional Training}

\vspace{-10pt}
\subsection{Language Instruction Following}
\vspace{-10pt}

Users primarily engage with generative AI models through natural language instructions, known as text prompts~\cite{brown2020language, zhou2022conditional}. 
%At the fine-tuning stage, the focus on instruction understanding and following 
Model instruction tuning aims to enhance AI models' capability to follow prompts accurately. This improved capability in prompt following enables models to generate output that more closely resembles human responses to natural language queries.
%effectively serve users' intended purpose, thereby aligning the output with user expectations. 
We start our discussion with a review of instruction following techniques for large language models (LLMs) and text-to-image models such as DALL·E 3. 
To enhance the text-to-video model's ability to follow text instructions, \texttt{Sora} utilizes an approach similar to that of DALL·E 3. The approach involves training a descriptive captioner and utilizing the captioner's generated data for fine-tuning. 
As a result of instruction tuning, %The ability to follow instructions refines 
\texttt{Sora} is able to accommodate a wide range of user requests, ensuring meticulous attention to the details in the instructions and generating videos that precisely meet users' needs.

\vspace{-10pt}
\subsubsection{Large Language Models}
\vspace{-10pt}

The capability of LLMs to follow instructions has been extensively explored~\cite{sanh2021multitask, wei2021finetuned, ouyang2022training}. This ability allows LLMs to read, understand, and respond appropriately to instructions describing an unseen task without examples. Prompt following ability is obtained and enhanced by fine-tuning LLMs on a mixture of tasks formatted as instructions\cite{sanh2021multitask, ouyang2022training}, known as instruction tuning. Wei et al.~\cite{wei2021finetuned} showed that instruction-tuned LLMs significantly outperform the untuned ones on unseen tasks. The instruction-following ability transforms LLMs into general-purpose task solvers, marking a paradigm shift in the history of AI development. %, moving from solving specific tasks to embracing user-defined tasks through improved generalization over unseen tasks. 
%Therefore, instruction tuning marks a paradigm shift, with LLMs now following instructions directly from user prompts rather than pre-determined tasks.

\vspace{-10pt}
\subsubsection{Text-to-Image}
\vspace{-10pt}

The instruction following in DALL·E 3 is addressed by a caption improvement method with a hypothesis that the quality of text-image pairs that the model is trained on determines the performance of the resultant text-to-image model~\cite{jia2021scaling}. The poor quality of data, particularly the prevalence of noisy data and short captions that omit a large amount of visual information, leads to many issues such as neglecting keywords and word order, and misunderstanding the user intentions~\cite{betker2023improving}. The caption improvement approach addresses these issues by re-captioning existing images with detailed, descriptive captions. The approach first trains an image captioner, which is a vision-language model, to generate precise and descriptive image captions. The resulting descriptive image captions by the captioner are then used to fine-tune text-to-image models. Specifically, DALL·E 3 follows contrastive captioners (CoCa)~\cite{yu2022coca} to jointly train an image captioner with a CLIP~\cite{radford2021learning} architecture and a language model objective. 
This image captioner incorporates an image encoder a unimodal text encoder for extracting language information, and a multimodal text decoder. It first employs a contrastive loss between unimodal image and text embeddings, followed by a captioning loss for the multimodal decoder's outputs. The resulting image captioner is further fine-tuned on a highly detailed description of images covering main objects, surroundings, backgrounds, texts, styles, and colorations. With this step, the image captioner is able to generate detailed descriptive captions for the images. The training dataset for the text-to-image model is a mixture of the re-captioned dataset generated by the image captioner and ground-truth human-written data to ensure that the model captures user inputs. This image caption improvement method introduces a potential issue: a mismatch between the actual user prompts and descriptive image descriptions from the training data. DALL·E 3 addresses this by \emph{upsampling}, where LLMs are used to re-write short user prompts into detailed and lengthy instructions. This ensures that the model's text inputs received in inference time are consistent with those in model training.

\vspace{-5pt}
\subsubsection{Text-to-Video}
\vspace{-5pt}

To enhance the ability of instruction following, \texttt{Sora} adopts a similar caption improvement approach. This method is achieved by first training a video captioner capable of producing detailed descriptions for videos. Then, this video captioner is applied to all videos in the training data to generate high-quality (video, descriptive caption) pairs, which are used to fine-tune \texttt{Sora} to improve its instruction following ability. 

\texttt{Sora}'s technical report~\cite{openai2024sora} does not reveal the details about how the video captioner is trained. Given that the video captioner is a video-to-text model, there are many approaches to building it. A straightforward approach is to utilize CoCa architecture for video captioning by taking multiple frames of a video and feeding each frame into the image encoder~\cite{yu2022coca}, known as VideoCoCa~\cite{yan2022video}. VideoCoCa builds upon CoCa and re-uses the image encoder pre-trained weights and applies it independently on sampled video frames. The resulting frame token embeddings are flattened and concatenated into a long sequence of video representations. These flattened frame tokens are then processed by a generative pooler and a contrastive pooler, which are jointly trained with the contrastive loss and captioning loss. Other alternatives to building video captioners include mPLUG-2~\cite{xu2023mplug}, GIT~\cite{wang2022git}, FrozenBiLM~\cite{yang2022zero}, and more. Finally, to ensure that user prompts align with the format of those descriptive captions in training data, \texttt{Sora} performs an additional prompt extension step, where GPT-4V is used to expand user inputs to detailed descriptive prompts.

\vspace{-5pt}
\subsubsection{Discussion}
\vspace{-5pt}

%The instruction-following ability of generative AI models is critical, as it involves training these models to understand and follow users' natural language instructions accurately. 
The instruction-following ability is critical for %more pronounced with the introduction of
\texttt{Sora} to generate one-minute-long videos with intricate scenes that are faithful to user intents. According to \texttt{Sora}'s technical report~\cite{openai2024sora}, this ability is obtained by developing a captioner that can generate long and detailed captions, which are then used to train the model. However, the process of collecting data for training such a captioner is unknown and likely labor-intensive, as it may require detailed descriptions of videos. Moreover, the descriptive video captioner might hallucinate important details of the videos. We believe that how to improve the video captioner warrants further investigation and is critical to enhance the instruction-following ability of text-to-image models.

\subsection{Prompt Engineering}
Prompt engineering refers to the process of designing and refining the input given to an AI system, particularly in the context of generative models, to achieve specific or optimized outputs~\cite{Li2023Apractical,chen2023unleashing,pitis2023boosted}. The art and science of prompt engineering involve crafting these inputs in a way that guides the model to produce the most accurate, relevant, and coherent responses possible.

\vspace{-10pt}
\subsubsection{Text Prompt}
\begin{figure}[h]
    \centering
    \includegraphics[width=\linewidth]{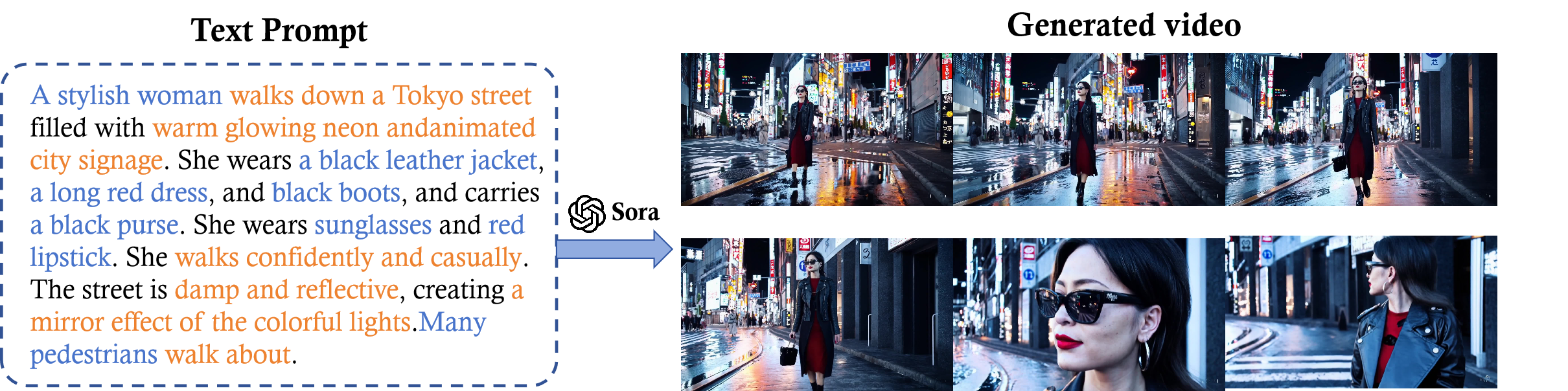}
    \caption{A case study on prompt engineering for text-to-video generation, employing color coding to delineate the creative process. The text highlighted in blue describes the elements generated by \texttt{Sora}, such as the depiction of a stylish woman. In contrast, the text in yellow accentuates the model's interpretation of actions, settings, and character appearances, demonstrating how a meticulously crafted prompt is transformed into a vivid and dynamic video narrative.}
    \label{fig: text prompt}
\end{figure}

\vspace{-10pt}
Text prompt engineering is vital in directing text-to-video models (e.g., \texttt{Sora} \cite{openai2024sora}) to produce videos that are visually striking while precisely meeting user specifications. This involves crafting detailed descriptions 
%that serve as instructions for 
to instruct the model to effectively bridge the gap between human creativity and AI's execution capabilities~\cite{hao2023optimizing}. The prompts for \texttt{Sora} cover a wide range of scenarios. Recent works (e.g., VoP~\cite{huang2023vop}, Make-A-Video~\cite{singer2022makeavideo}, and Tune-A-Video~\cite{wu2023tuneavideo}) have shown how prompt engineering leverages model's natural language understanding ability to decode complex instructions and render them into cohesive, lively, and high-quality video narratives. As shown in Figure~\ref{fig: text prompt}, ``a stylish woman walking down a neon-lit Tokyo street...'' is such a meticulously crafted text prompt that it ensures \texttt{Sora} to generate a video that aligns well with the expected vision. The quality of prompt engineering depends on the careful selection of words, the specificity of the details provided, and comprehension of their impact on the model's output. For example, the prompt in Figure~\ref{fig: text prompt} specifies in detail the actions, settings, character appearances, and even the desired mood and atmosphere of the scene. 
%These instructions instruct the model what to produce and shape the narrative's dynamics, encompassing the visual aesthetics, character movement, and object interactions.

\vspace{-10pt}
\subsubsection{Image Prompt}
\begin{figure}[h]
    \centering
    \includegraphics[width=\linewidth]{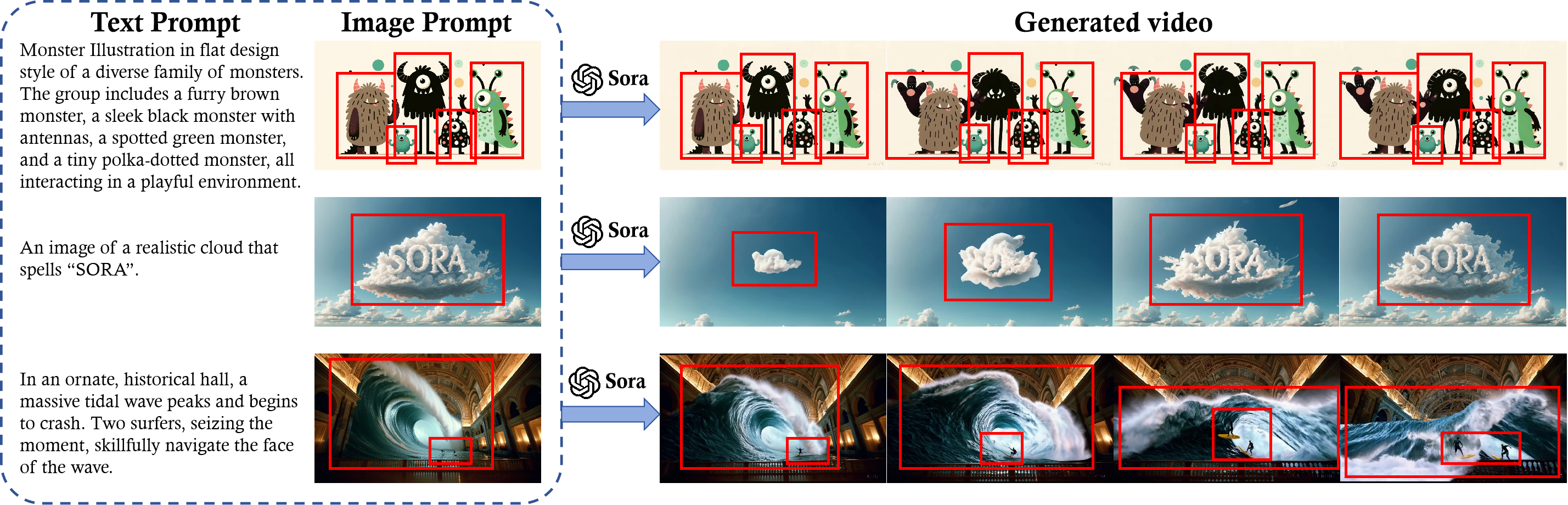}
    \vspace{-20pt}
    \caption{This example illustrates the image prompts to guide \texttt{Sora}'s text-to-video model to generation. The red boxes visually anchor the key elements of each scene—monsters of varied designs, a cloud formation spelling ``SORA'', and surfers in an ornate hall facing a massive tidal wave.}
    \label{fig: image prompt}
    \vspace{-10pt}
\end{figure}

\vspace{-10pt}
An image prompt serves as a visual anchor for the to-be-generated video's content and other elements such as
%establishing the groundwork for its 
characters, setting, and mood \cite{Luddecke2022image}. 
In addition, a text prompt can instruct the model to animate these elements by e.g., adding layers of movement, interaction, and narrative progression that bring the static image to life ~\cite{blattmann2023stable,chen2023seine,chen2024videocrafter2}.
%, as Stable Video Diffusion~\cite{blattmann2023stable}, SEING~\cite{chen2023seine} and VideoCrafter2~\cite{chen2024videocrafter2}. 
The use of image prompts %for \texttt{Sora} 
%represents a significant advancement in the field of AI-generated video content, which 
allows \texttt{Sora} to convert static images into dynamic, narrative-driven videos by leveraging both visual and textual information. 
In Figure~\ref{fig: image prompt}, we show AI-generated videos of ``a Shiba Inu wearing a beret and turtleneck'', ``a unique monster family'', ``a cloud forming the word `{SORA}''' and ``surfers navigating a tidal wave inside a historic hall''. These examples demonstrate what can be achieved by prompting \texttt{Sora} with DALL·E-generated images.
%the broad and imaginative possibilities of integrating DALL·E-generated images with \texttt{Sora}'s video generation technology.

% The key to effective image prompt engineering lies in the seamless fusion of visual and textual information.

\vspace{-10pt}
\subsubsection{Video Prompt}
\vspace{-10pt}

\begin{figure}[h]
    \centering
    \includegraphics[width=\linewidth]{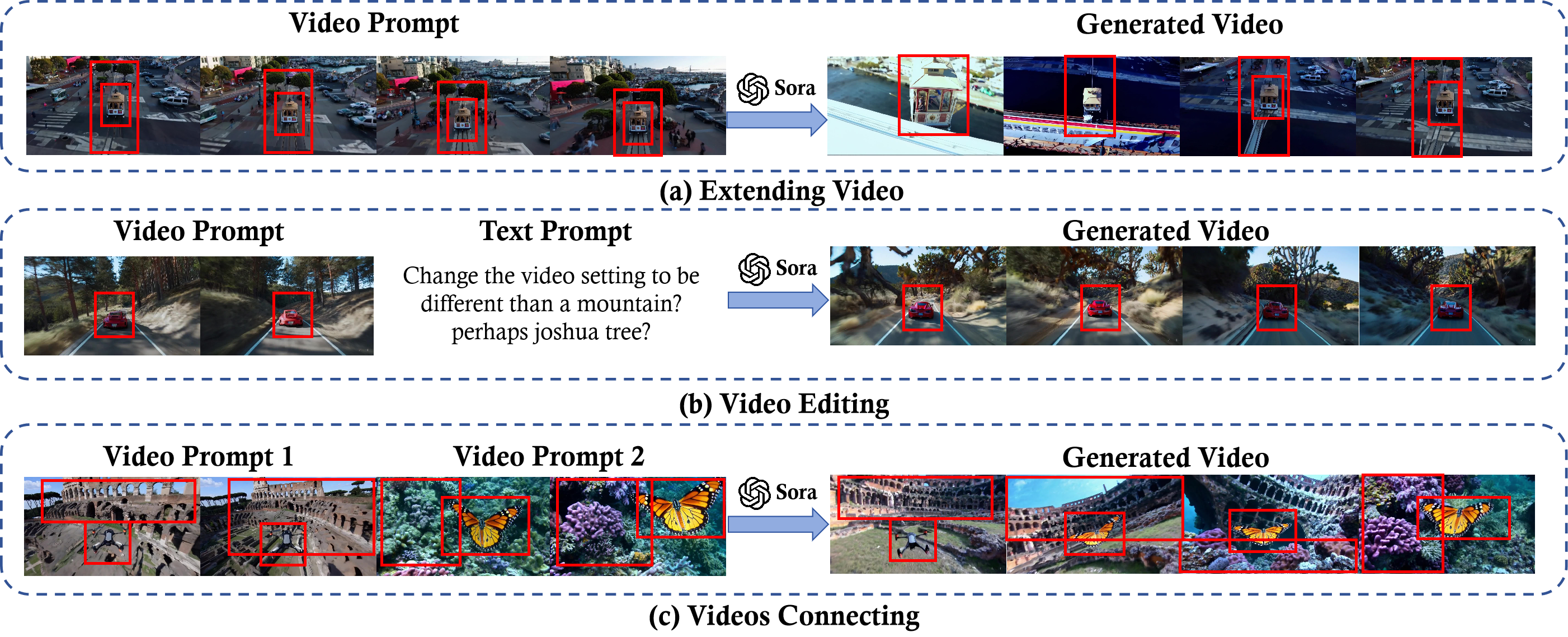}
    \vspace{-20pt}
    \caption{These examples illustrate the video prompt techniques for \texttt{Sora} models: (a) Video Extension, where the model extrapolates the sequence backward the original footage, (b) Video Editing, where specific elements like the setting are transformed as per the text prompt, and (c) Video Connection, where two distinct video prompts are seamlessly blended to create a coherent narrative. Each process is guided by a visual anchor, marked by a red box, ensuring continuity and precision in the generated video content.}
    \label{fig: video prompt}
\end{figure}

Video prompts can also be used for video generation
%acts as a guide for generative models, instructing them on how to create, develop, or modify videos in a manner that meets the user's goals, 
as demonstrated in ~\cite{wang2018videotovideo,wang2019fewshot}. 
Recent works (e.g., Moonshot~\cite{zhang2024moonshot} and Fast-Vid2Vid~\cite{zhuo2022fastvid2vid}) show that good video prompts need to be specific and flexible.
%require a balance between specificity and flexibility. 
This ensures that the model receives clear direction on specific objectives, like the portrayal of particular objects and visual themes, and also allows for imaginative variations in the final output. For example, in the video extension tasks, a prompt could specify the direction (forward or backward in time) and the context or theme of the extension. 
In Figure~\ref{fig: video prompt}(a), the video prompt instructs \texttt{Sora} to extend a video backward in time to explore the events leading up to the original starting point. 
When performing video-to-video editing through video prompts, as shown in Figure~\ref{fig: video prompt}(b), the model needs to clearly understand the desired transformation, such as changing the video's style, setting or atmosphere, or altering subtle aspects like lighting or mood. 
%In the context of connecting videos, as shown 
In Figure~\ref{fig: video prompt}(c), the prompt instructs \texttt{Sora} to connect videos while ensuring smooth transitions between objects in different scenes across videos.

\vspace{-5pt}

\vspace{-10pt}
\subsubsection{Discussion}
\vspace{-10pt}

Prompt engineering %is a critical interface between human intent and AI capabilities, enabling
allows users to guide AI models to generate content that aligns with their intent. As an example, the combined use of text, image, and video prompts enables \texttt{Sora} to create content that is not only visually compelling but also aligned well with users' expectations and intent. 
%Historically, research on making AI model to understand and follow human instructions has mainly focused on LLMs or LVMs~\cite{liu2021pretrain,lester2021power,jia2022visual}, with a focus on generating text or static images.
While previous studies on prompt engineering have been focused on text and image prompts for LLMs and LVMs~\cite{liu2021pretrain,lester2021power,jia2022visual},
we expect that there will be a growing interest in video prompts for video generation models.
%However, the study of how to best guide AI in creating videos from text prompts—especially in making these prompts more effective for video generation—is not as developed, which will be a future direction of research.

\vspace{-10pt}
\subsection{Trustworthiness}
\vspace{-10pt}

With the rapid advancement of sophisticated models such as ChatGPT \cite{chatgpt}, GPT4-V \cite{gpt4v}, and \texttt{Sora} \cite{openai2024sora}, the capabilities of these models have seen remarkable enhancements. These developments have made significant contributions to improving work efficiency and propelling technological progress. However, these advancements also raise concerns about the potential for misuse of these technologies, including the generation of fake news \cite{huang2023harnessing, chen2023llmgenerated}, privacy breaches \cite{liu2023deidgpt}, and ethical dilemmas \cite{yao2023value, huang2023trustgpt}. Consequently, the issue of trustworthiness in large models has garnered extensive attention from both the academic and industrial spheres, emerging as a focal point of contemporary research discussions.

\vspace{-10pt}
\subsubsection{Safety Concern}
\vspace{-10pt}

One primary area of focus is the model's safety, specifically its resilience against misuse and so-called ``jailbreak'' attacks, where users attempt to exploit vulnerabilities to generate prohibited or harmful content \cite{sun2024trustllm, mazeika2024harmbench, wang2023donotanswer, wang2023decodingtrust, zhang2023safetybench, shen2023anything, liu2023autodan, zhu2023autodan, zhou2024robust, guo2024coldattack}. For instance, AutoDAN \cite{zhu2023autodan}, a novel and interpretable adversarial attack method based on gradient techniques, is introduced to enable system bypass. In a recent study, researchers explore two reasons why LLMs struggle to resist jailbreak attacks: competing objectives and mismatched generalization \cite{wei2023jailbroken}. Besides textual attacks, visual jailbreak also threatens the safety of multimodal models (e.g., GPT-4V \cite{gpt4v}, and \texttt{Sora} \cite{openai2024sora}). A recent study  \cite{niu2024jailbreaking} found that large multimodal models are more vulnerable since the continuous and high-dimensional nature of the additional visual input makes it weaker against adversarial attacks, representing an expanded attack surface.

\vspace{-10pt}
\subsubsection{Other Exploitation}
\vspace{-10pt}

% Another vital aspect of evaluation is fairness and bias. The importance of developing models that do not amplify societal biases cannot be overstated. Studies in this area \cite{gallegos2023bias, zhang2023chatgpt, liang2023detection, friedrich2023fair} aim to uncover and address inherent biases within models, ensuring that AI systems are equitable and do not discriminate based on race, gender, or other sensitive attributes. Privacy preservation also stands out as a cornerstone of trustworthy AI. With increasing awareness and concern over data privacy, evaluations now rigorously assess how well models safeguard user data and ensure that personal information is not inadvertently revealed \cite{mireshghallah2023can, plant2022you, li2023privacy}. 

Due to the large scale of the training dataset and training methodology of large foundation models (e.g., ChatGPT \cite{chatgpt} and \texttt{Sora} \cite{openai2024sora}), the truthfulness of these models needs to be enhanced as the related issues like hallucination have been discussed widely \cite{liu2024survey}. Hallucination in this context refers to the models' tendency to generate responses that may appear convincing but are unfounded or false \cite{sun2024trustllm}. This phenomenon raises critical questions about the reliability and trustworthiness of model outputs, necessitating a comprehensive approach to both evaluate and address the issue. Amount of studies have been dedicated to dissecting the problem of hallucination from various angles. This includes efforts aimed at evaluating the extent and nature of hallucination across different models and scenarios \cite{guan2023hallusionbench, sun2024trustllm, li2023evaluating, huang2023metatool}. These evaluations provide invaluable insights into how and why hallucinations occur, laying the groundwork for developing strategies to mitigate their incidence. Concurrently, a significant body of research is focused on devising and implementing methods to reduce hallucinations in these large models \cite{liu2023mitigating, wang2024mitigating, zhou2023analyzing}. 

\noindent Another vital aspect of trustworthiness is fairness and bias. The critical importance of developing models that do not perpetuate or exacerbate societal biases is a paramount concern. This priority stems from the recognition that biases encoded within these models can reinforce existing social inequities, leading to discriminatory outcomes. Studies in this area, as evidenced by the work of Gallegos et al. \cite{gallegos2023bias}, Zhang et al. \cite{zhang2023chatgpt}, Liang et al. \cite{liang2023detection}, and Friedrich et al. \cite{friedrich2023fair}, are dedicated to the meticulous identification and rectification of these inherent biases. The goal is to cultivate models that operate fairly, treating all individuals equitably without bias towards race, gender, or other sensitive attributes. This involves not only detecting and mitigating bias in datasets but also designing algorithms that can actively counteract the propagation of such biases \cite{Liu_Jia_Wei_Xu_Wang_Vosoughi_2021, mahabadi2020endtoend}.

\noindent Privacy preservation emerges as another foundational pillar when these models are deployed. In an era where data privacy concerns are escalating, the emphasis on protecting user data has never been more critical. The increasing public awareness and concern over how personal data is handled have prompted more rigorous evaluations of large models. These evaluations focus on the models' capacity to protect user data, ensuring that personal information remains confidential and is not inadvertently disclosed. Research by Mireshghallah et al. \cite{mireshghallah2023can}, Plant et al. \cite{plant2022you}, and Li et al. \cite{li2023privacy} exemplify efforts to advance methodologies and technologies that safeguard privacy.

\vspace{-10pt}
\subsubsection{Alignment}
\vspace{-10pt}

In addressing these challenges, ensuring the trustworthiness of large models has become one of the primary concerns for researchers \cite{bommasani2022opportunities, sun2024trustllm, wang2023decodingtrust, shen2023large}. Among the most important technologies is model alignment \cite{shen2023large, liu2023alignbench}, which refers to {the process and goal of ensuring that the behavior and outputs of models are consistent with the intentions and ethical standards of human designers}. This concerns the development of technology, its moral responsibilities, and social values. In the domain of LLMs, the method of Reinforcement Learning with Human Feedback (RLHF) \cite{christiano2023deep, yu2023rlhfv} has been widely applied for model alignment. This method combines Reinforcement Learning (RL) with direct human feedback, allowing models to better align with human expectations and standards in understanding and performing tasks.

\vspace{-10pt}
\subsubsection{Discussion}
\vspace{-10pt}

From \texttt{Sora} (specifically its technical report), we summarize some insightful findings that potentially offer an informative guideline for future work:

\noindent(1) \textit{Integrated Protection of Model and External Security}: As models become more powerful, especially in generating content, ensuring that they are not misused to produce harmful content (such as hate speech \cite{jahan2023systematic} and false information \cite{chen2023llmgenerated, huang2023harnessing}) has become a serious challenge. In addition to aligning the model itself, external security protections are equally important. This includes content filtering and review mechanisms, usage permissions and access control, data privacy protection, as well as enhancements in transparency and explainability. For instance, OpenAI now uses a detection classifier to tell whether a given video is generated by \texttt{Sora} \cite{openai2024sorasafety}. Moreover, a text classifier is deployed to detect the potentially harmful textual input \cite{openai2024sorasafety}. 

\noindent(2) \textit{Security Challenges of Multimodal Models}: Multimodal models, such as text-to-video models like \texttt{Sora} bring additional complexity to security due to their ability to understand and generate various types of content (text, images, videos, etc.). Multimodal models can produce content in various forms, increasing the ways and scope of misuse and copyright issues. As the content generated by multimodal models is more complex and diverse, traditional methods of content verification and authenticity may no longer be effective. This requires the development of new technologies and methods to identify and filter harmful content generated by these models, increasing the difficulty of regulation and management. 

\noindent(3) \textit{The Need for Interdisciplinary Collaboration}: Ensuring the safety of models is not just a technical issue but also requires cross-disciplinary cooperation. To address these challenges, experts from various fields such as law \cite{fei2023lawbench} and psychology \cite{li2024i} need to work together to develop appropriate norms (e.g., what's the safety and what's unsafe?), policies, and technological solutions. The need for interdisciplinary collaboration significantly increases the complexity of solving these issues.

\vspace{-10pt}
\section{Applications}
\vspace{-10pt}

\begin{figure}[t]
    \centering
    \includegraphics[width=0.9\linewidth]{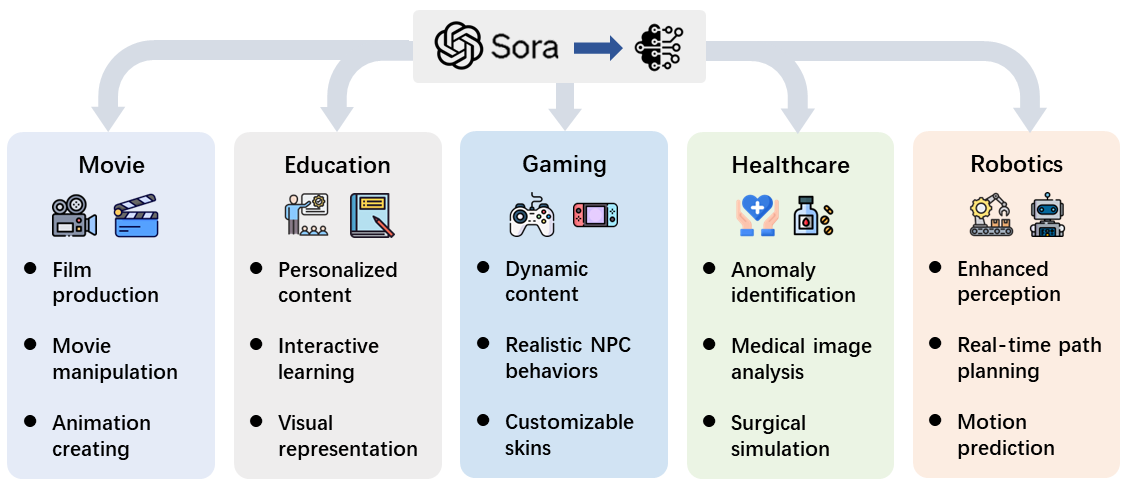}
    \caption{Applications of \texttt{Sora}.}
    \label{fig: application}
\end{figure}

As video diffusion models, exemplified by \texttt{Sora}, emerge as a forefront technology, their adoption across diverse research fields and industries is rapidly accelerating. The implications of this technology extend far beyond mere video creation, offering transformative potential for tasks ranging from automated content generation to complex decision-making processes. In this section, we delve into a comprehensive examination of the current applications of video diffusion models, highlighting key areas where \texttt{Sora} has not only demonstrated its capabilities but also revolutionized the approach to solving complex problems. We aim to offer a broad perspective for the practical deployment scenarios (see Figure \ref{fig: application}).

\vspace{-10pt}
\subsection{Movie}
\vspace{-10pt}

Traditionally, creating cinematic masterpieces has been an arduous and expensive process, often requiring decades of effort, cutting-edge equipment, and substantial financial investments. However, the advent of advanced video generation technologies heralds a new era in film-making, one where the dream of autonomously producing movies from simple text inputs is becoming a reality. Researchers have ventured into the realm of movie generation by extending video generation models into creating movies. MovieFactory~\cite{zhu2023moviefactory} applies diffusion models to generate film-style videos from elaborate scripts produced by ChatGPT \cite{chatgpt}, representing a significant leap forward. In the follow-up, MobileVidFactory~\cite{zhu2023mobilevidfactory} can automatically generate vertical mobile videos with only simple texts provided by users. Vlogger~\cite{zhuang2024vlogger} makes it feasible for users to compose a minute-long vlog. These developments, epitomized by \texttt{Sora}'s ability to generate captivating movie content effortlessly, mark a pivotal moment in the democratization of movie production. They offer a glimpse into a future where anyone can be a filmmaker, significantly lowering the barriers to entry in the film industry and introducing a novel dimension to movie production that blends traditional storytelling with AI-driven creativity. The implications of these technologies extend beyond simplification. They promise to reshape the landscape of film production, making it more accessible and versatile in the face of evolving viewer preferences and distribution channels.

\vspace{-10pt}
\subsection{Education}
\vspace{-10pt}

The landscape of educational content has long been dominated by static resources, which, despite their value, often fall short of catering to the diverse needs and learning styles of today’s students. Video diffusion models stand at the forefront of an educational revolution, offering unprecedented opportunities to customize and animate educational materials in ways that significantly enhance learner engagement and understanding. These advanced technologies enable educators to transform text descriptions or curriculum outlines into dynamic, engaging video content tailored to the specific style, and interests of individual learners~\cite{feng2023ccedit,xing2023make,guo2023animatediff,he2023animate}. Moreover, image-to-video editing techniques~\cite{ni2023conditional, hu2023animate,hu2022make} present innovative avenues for converting static educational assets into interactive videos, thereby supporting a range of learning preferences and potentially increasing student engagement. By integrating these models into educational content creation, educators can produce videos on a myriad of subjects, making complex concepts more accessible and captivating for students. The use of \texttt{Sora} in revolutionizing the educational domain exemplifies the transformative potential of these technologies. This shift towards personalized, dynamic educational content heralds a new era in education.

\vspace{-10pt}
\subsection{Gaming}
\vspace{-10pt}

The gaming industry constantly seeks ways to push the boundaries of realism and immersion, yet traditional game development often grapples with the limitations of pre-rendered environments and scripted events. The generation of dynamic, high-fidelity video content and realistic sound by diffusion models effects in real-time, promise to overcome existing constraints, offering developers the tools to create evolving game environments that respond organically to player actions and game events~\cite{mei2023vidm,yu2023video}. This could include generating changing weather conditions, transforming landscapes, or even creating entirely new settings on the fly, making game worlds more immersive and responsive.
Some methods~\cite{su2023physics,li2024dance} also synthesize realistic impact sounds from video inputs, enhancing game audio experiences. With the integration of \texttt{Sora} within the gaming domain, unparalleled immersive experiences that captivate and engage players can be created. How games are developed, played, and experienced will be innovated, as well as opening new possibilities for storytelling, interaction, and immersion.

\vspace{-10pt}
\subsection{Healthcare}
\vspace{-10pt}

Despite generative capabilities, video diffusion models excel in understanding and generating complex video sequences, making them particularly suited for identifying dynamic anomalies within the body, such as early cellular apoptosis~\cite{awasthi2023video}, skin lesion progression~\cite{bozorgpour2023dermosegdiff}, and irregular human movements~\cite{flaborea2023multimodal}, which are crucial for early disease detection and intervention strategies. Additionally, models like MedSegDiff-V2~\cite{wu2023medsegdiff} and \cite{chowdary2023diffusion} leverage the power of transformers to segment medical images with unprecedented precision, enabling clinicians to pinpoint areas of interest across various imaging modalities with enhanced accuracy.
The integration of \texttt{Sora} into clinical practice promises not only to refine diagnostic processes but also to personalize patient care, offering tailored treatment plans based on precise medical imaging analysis. However, this technological integration comes with its own set of challenges, including the need for robust data privacy measures and addressing ethical considerations in healthcare.

\vspace{-10pt}
\subsection{Robotics}
\vspace{-10pt}

Video diffusion models now play important roles in robotics, showing a new era where robots can generate and interpret complex video sequences for enhanced perception~\cite{kapelyukh2023dall,liu2022structdiffusion} and decision-making~\cite{janner2022planning,ajay2022conditional,carvalho2023motion}. These models unlock new capabilities in robots, enabling them to interact with their environment and execute tasks with unprecedented complexity and precision. The introduction of web-scale diffusion models to robotics~\cite{kapelyukh2023dall} showcases the potential for leveraging large-scale models to enhance robotic vision and understanding. Latent diffusion models are employed for language-instructed video prediction~\cite{gu2023seer}, allowing robots to understand and execute tasks by predicting the outcome of actions in video format. 
Furthermore, the reliance on simulated environments for robotics research has been innovatively addressed by video diffusion models capable of creating highly realistic video sequences~\cite{chen2023genaug,mandi2022cacti}. This enables the generation of diverse training scenarios for robots, mitigating the limitations imposed by the scarcity of real-world data.
We believe, the integration of technologies like \texttt{Sora} into the robotics field holds the promise of groundbreaking developments. By harnessing the power of \texttt{Sora}, the future of robotics is poised for unprecedented advancements, where robots can seamlessly navigate and interact with their environments. 

\vspace{-10pt}
\section{Discussion}
\vspace{-10pt}

\texttt{Sora} shows a remarkable talent for precisely understanding and implementing complex instructions from humans. This model excels at creating detailed videos with various characters, all set within elaborately crafted settings. A particularly impressive attribute of \texttt{Sora} is its ability to produce videos up to one minute in length while ensuring consistent and engaging storytelling. This marks a significant improvement over previous attempts that focused on shorter video pieces, as \texttt{Sora}'s extended sequences exhibit a clear narrative flow and maintain visual consistency from start to finish. Furthermore, \texttt{Sora} distinguishes itself by generating longer video sequences that capture complex movements and interactions, advancing past the restrictions of earlier models that could only handle short clips and basic images. This advancement signifies a major step forward in AI-powered creative tools, enabling users to transform written stories into vivid videos with a level of detail and sophistication that was previously unattainable.

\vspace{-10pt}
\subsection{Limitations}
\vspace{-10pt}

\hhead{Challenges in Physical Realism.} \texttt{Sora}, as a simulation platform, exhibits a range of limitations that undermine its effectiveness in accurately depicting complex scenarios. Most important is its inconsistent handling of physical principles within complex scenes, leading to a failure in accurately copying specific examples of cause and effect. For instance, consuming a portion of a cookie might not result in a corresponding bite mark, illustrating the system's occasional departure from physical plausibility. This issue extends to the simulation of motion, where \texttt{Sora} generates movements that challenge realistic physical modeling, such as unnatural transformations of objects or the incorrect simulation of rigid structures like chairs, leading to unrealistic physical interactions. The challenge further increases when simulating complex interactions among objects and characters, occasionally producing outcomes that lean towards the humorous.

\hhead{Spatial and Temporal Complexities.} \texttt{Sora} occasionally misunderstands instructions related to the placement or arrangement of objects and characters within a given prompt, leading to confusion about directions (e.g., confusing left for right). Additionally, it faces challenges in maintaining the temporal accuracy of events, particularly when it comes to adhering to designated camera movements or sequences. This can result in deviating from the intended temporal flow of scenes. In complex scenarios that involve a multitude of characters or elements, \texttt{Sora} has a tendency to insert irrelevant animals or people. Such additions can significantly change the originally envisioned composition and atmosphere of the scene, moving away from the planned narrative or visual layout. This issue not only affects the model's ability to accurately recreate specific scenes or narratives but also impacts its reliability in generating content that closely aligns with the user's expectations and the coherence of the generated output. 

\hhead{Limitations in Human-computer Interaction (HCI).} \texttt{Sora}, while showing potential in the video generation domain, faces significant limitations in HCI. These limitations are primarily evident in the coherence and efficiency of user-system interactions, especially when making detailed modifications or optimizations to generated content. For instance, users might find it difficult to precisely specify or adjust the presentation of specific elements within a video, such as action details and scene transitions. Additionally, \texttt{Sora}'s limitations in understanding complex language instructions or capturing subtle semantic differences could result in video content that does not fully meet user expectations or needs. These shortcomings restrict \texttt{Sora}'s potential in video editing and enhancement, also impacting the overall satisfaction of the user experience.

\hhead{Usage Limitation.} 
Regarding usage limitations, OpenAI has not yet set a specific release date for public access to \texttt{Sora}, emphasizing a cautious approach towards safety and readiness before broad deployment. This indicates that further improvements and testing in areas such as security, privacy protection, and content review may still be necessary for \texttt{Sora}. Moreover, at present, \texttt{Sora} can only generate videos up to one minute in length, and according to published cases, most generated videos are only a few dozen seconds long. This limitation restricts its use in applications requiring longer content display, such as detailed instructional videos or in-depth storytelling. This limitation reduces \texttt{Sora}'s flexibility in the content creation.

\vspace{-10pt}
\subsection{Opportunities}
\vspace{-10pt}

\hhead{Academy.}
(1) The introduction of \texttt{Sora} by OpenAI marks a strategic shift towards encouraging the broader AI community to delve deeper into the exploration of text-to-video models, leveraging both diffusion and transformer technologies. This initiative aims to redirect the focus toward the potential of creating highly sophisticated and nuanced video content directly from textual descriptions, a frontier that promises to revolutionize content creation, storytelling, and information sharing. (2) The innovative approach of training \texttt{Sora} on data at its native size, as opposed to the traditional methods of resizing or cropping, serves as a groundbreaking inspiration for the academic community. It opens up new pathways by highlighting the benefits of utilizing unmodified datasets, which leads to the creation of more advanced generative models.

\hhead{Industry.}
(1) The current capabilities of \texttt{Sora} signal a promising path for the advancement of video simulation technologies, highlighting the potential to significantly enhance realism within both physical and digital areas. The prospect of \texttt{Sora} enabling the creation of highly realistic environments through textual descriptions presents a promising future for content creation. This potential extends to revolutionizing game development, offering a glimpse into a future where immersive-generated worlds can be crafted with unprecedented ease and accuracy. (2) Companies may leverage \texttt{Sora} to produce advertising videos that swiftly adapt to market changes and create customized marketing content. This not only reduces production costs but also enhances the appeal and effectiveness of advertisements. The ability of \texttt{Sora} to generate highly realistic video content from textual descriptions alone could revolutionize how brands engage with their audience, allowing for the creation of immersive and compelling videos that capture the essence of their products or services in unprecedented ways.

\hhead{Society.}
(1) While the prospect of utilizing text-to-video technology to replace traditional filmmaking remains distant, \texttt{Sora} and similar platforms hold transformative potential for content creation on social media. The constraints of current video lengths do not diminish the impact these tools can have in making high-quality video production accessible to everyone, enabling individuals to produce compelling content without the need for expensive equipment. It represents a significant shift towards empowering content creators across platforms like TikTok and Reels, bringing in a new age of creativity and engagement. (2) Screenwriters and creative professionals can use \texttt{Sora} to transform written scripts into videos, assisting them in better showing and sharing their creative concepts, and even in producing short films and animations. The ability to create detailed, vivid videos from scripts can fundamentally change the pre-production process of filmmaking and animation, offering a glimpse into how future storytellers might pitch, develop, and refine their narratives. This technology opens up possibilities for a more dynamic and interactive form of script development, where ideas can be visualized and assessed in real time, providing a powerful tool for creativity and collaboration. (3) Journalists and news organizations can also utilize \texttt{Sora} to quickly generate news reports or explanatory videos, making the news content more vivid and engaging. This can significantly increase the coverage and audience engagement of news reports. By providing a tool that can simulate realistic environments and scenarios, \texttt{Sora} offers a powerful solution for visual storytelling, enabling journalists to convey complex stories through engaging videos that were previously difficult or expensive to produce. In summary, \texttt{Sora}'s potential to revolutionize content creation across marketing, journalism, and entertainment is immense.

\vspace{-10pt}
\section{Conclusion}
\vspace{-10pt}

We present a comprehensive review of \texttt{Sora} to help developers and researchers study the capabilities and related works of \texttt{Sora}.
The review is based on our survey of published technical reports and reverse engineering based on existing literature.
We will continue to update the paper when \texttt{Sora}'s API is available and further details about \texttt{Sora} are revealed.
%Based on better understandings of \texttt{Sora},
We hope that this review paper will prove a valuable resource for the open-source research community and lay a foundation for the community to jointly develop an open-source version of \texttt{Sora} in the near future to democratize video auto-creation in the era of AIGC. To achieve this goal, we invite discussions, suggestions, and collaborations on all fronts. 

\bibliographystyle{ieeetr}
\bibliography{LVM}

\pagebreak
\appendix
\section{Related Works}
We show some related works about the video generation tasks in Table \ref{tab:LVM for video gen}.

\begin{table}[!htbp]
	% \tiny
	\centering
	\caption{Summary of Video Generation.}
	\label{tab:LVM for video gen}
	\resizebox{0.9\textwidth}{!}{
	\begin{threeparttable}
	%\resizebox{\textwidth}{!}{
	%\begin{tabular}{p{0.2cm}p{0.6cm}p{1.4cm}p{1.7cm}p{1.6cm}p{1.8cm}p{2.8cm}l}
    \begin{tabular}{lllll}\hline
            \textbf{Model name}    & \textbf{Year} & \textbf{Backbone}    & \textbf{Task}       & \textbf{Group}                 \\ \hline
Imagen Video\cite{ho2022imagen}           & 2022          & Diffusion            & Generation          & Google                         \\
Pix2Seq-D\cite{chen2023generalist}              & 2022          & Diffusion            & Segmentation        & Google Deepmind                \\
FDM\cite{harvey2022flexible}                    & 2022          & Diffusion            & Prediction          & UBC                            \\
MaskViT\cite{gupta2022maskvit}                & 2022          & Masked Vision Models & Prediction          & Stanford, Salesforce           \\
CogVideo\cite{hong2022cogvideo}               & 2022          & Auto-regressive      & Generation          & THU                            \\
Make-a-video\cite{singer2022make}           & 2022          & Diffusion            & Generation          & Meta                           \\
MagicVideo\cite{zhou2022magicvideo}             & 2022          & Diffusion            & Generation          & ByteDance                      \\
TATS\cite{ge2022long}                   & 2022          & Auto-regressive      & Generation          & University of Maryland, Meta   \\
Phenaki\cite{villegas2022phenaki}                & 2022          & Masked Vision Models & Generation          & Google Brain                   \\
Gen-1\cite{esser2023structure}                 & 2023          & Diffusion            & Generation, Editing & RunwayML                       \\
LFDM\cite{ni2023conditional}                   & 2023          & Diffusion            & Generation          & PSU, UCSD                      \\
Text2video-Zero\cite{khachatryan2023text2video}        & 2023          & Diffusion            & Generation          & Picsart                        \\
Video Fusion\cite{luo2023videofusion}           & 2023          & Diffusion            & Generation          & USAC, Alibaba                  \\
PYoCo\cite{ge2023preserve}                  & 2023          & Diffusion            & Generation          & Nvidia                         \\
Video LDM\cite{blattmann2023align}              & 2023          & Diffusion            & Generation          & University of Maryland, Nvidia \\
RIN\cite{jabri2022scalable}                    & 2023          & Diffusion            & Generation          & Google Brain                   \\
LVD\cite{lian2023llm}                    & 2023          & Diffusion            & Generation          & UCB                            \\
Dreamix\cite{molad2023dreamix}                & 2023          & Diffusion            & Editing             & Google                         \\
MagicEdit\cite{liew2023magicedit}              & 2023          & Diffusion            & Editing             & ByteDance                      \\
Control-A-Video\cite{chen2023control}        & 2023          & Diffusion            & Editing             & Sun Yat-Sen University         \\
StableVideo\cite{chai2023stablevideo}            & 2023          & Diffusion            & Editing             & ZJU, MSRA                      \\
Tune-A-Video\cite{wu2023tuneavideo}           & 2023          & Diffusion            & Editing             & NUS                            \\
Rerender-A-Video\cite{yang2023rerender}       & 2023          & Diffusion            & Editing             & NTU                            \\
Pix2Video\cite{ceylan2023pix2video}              & 2023          & Diffusion            & Editing             & Adobe, UCL                     \\
InstructVid2Vid\cite{qin2023instructvid2vid}        & 2023          & Diffusion            & Editing             & ZJU                            \\
DiffAct\cite{liu2023diffusion}                & 2023          & Diffusion            & Action Detection    & University of Sydney           \\
DiffPose\cite{feng2023diffpose}               & 2023          & Diffusion            & Pose Estimation     & Jilin University               \\
MAGVIT\cite{yu2023magvit}                & 2023          & Masked Vision Models & Generation          & Google                         \\
AnimateDiff\cite{guo2023animatediff}            & 2023          & Diffusion            & Generation          & CUHK                           \\
MAGVIT V2\cite{yu2023language}              & 2023          & Masked Vision Models & Generation          & Google                         \\
Generative Dynamics\cite{li2023generative}    & 2023          & Diffusion            & Generation          & Google                         \\
VideoCrafter\cite{chen2024videocrafter2}           & 2023          & Diffusion            & Generation          & Tencent                        \\
Zeroscope\cite{Zeroscope}              & 2023          & -                    & Generation          & EasyWithAI                     \\
ModelScope             & 2023          & -                    & Generation          & Damo                           \\
Gen-2\cite{gen22023}                  & 2023          & -                    & Generation          & RunwayML                       \\
Pika\cite{pika2023}                  & 2023          & -                    & Generation          & Pika Labs                      \\
Emu Video\cite{girdhar2023emu}              & 2023          & Diffusion            & Generation          & Meta                           \\
PixelDance\cite{zeng2023make}             & 2023          & Diffusion            & Generation          & ByteDance                      \\
Stable Video Diffusion\cite{blattmann2023stable} & 2023          & Diffusion            & Generation          & Stability AI                   \\
W.A.L.T\cite{gupta2023photorealistic}                & 2023          & Diffusion            & Generation          & Stanford, Google               \\
Fairy\cite{wu2023fairy}                  & 2023          & Diffusion            & Generation, Editing & Meta                           \\
VideoPoet\cite{kondratyuk2023videopoet}              & 2023          & Auto-regressive      & Generation, Editing & Google                         \\
GenTron \cite{chen2023gentron} & 2023 & Diffusion & Generation & HKU \& Meta \\
LGVI\cite{wu2024towards}                   & 2024          & Diffusion       & Editing             & PKU, NTU                       \\
Lumiere\cite{bar2024lumiere}                & 2024          & Diffusion            & Generation          & Google                         \\
Sora\cite{openai2024sora}                   & 2024          & Diffusion            & Generation, Editing & OpenAI  \\ \bottomrule
	\end{tabular}
	% \begin{tablenotes}
	%     \tiny
	%     \item[1] Downstream task types: classification (cla), recognition (rec), detection (det), localization (loc), segmentation (seg), clustering (clu), inpainting (inp), retrieval (ret), generation (gen), pose estimation (pos), reinforcement learning (rel).
	% \end{tablenotes}
	\end{threeparttable}
	}
\end{table}

\end{document}